\documentclass[lettersize,journal]{IEEEtran}
\usepackage{amsmath,amsfonts}
\usepackage{algorithmic}
\usepackage{algorithm}
\usepackage{array}
\usepackage[caption=false,font=normalsize,labelfont=sf,textfont=sf]{subfig}
\usepackage{textcomp}
\usepackage{stfloats}
\usepackage{url}
\usepackage{verbatim}
\usepackage{graphicx}
\usepackage{cite}
\usepackage{hyperref}
\DeclareMathOperator*{\argmin}{arg\,min}
\usepackage[table,xcdraw]{xcolor}
\usepackage{booktabs}
\usepackage{graphicx}
\usepackage{multirow}
\usepackage{xcolor}
\usepackage{colortbl}
\usepackage{tikz}

\definecolor{tablerowgray}{RGB}{229, 229, 229}

\newlength{\squaresize}
\newlength{\squareborder}
\newlength{\squarespacing}
\newlength{\squarebaseline}
\newlength{\circlesize}
\definecolor{circletextcolor}{RGB}{255, 255, 255}  %

\newcommand{\circlefontsize}{\small\bfseries}  

\definecolor{rank1color}{RGB}{255, 0, 0}     %
\definecolor{rank2color}{RGB}{244, 0, 10}     %
\definecolor{rank3color}{RGB}{232, 0, 21}     %
\definecolor{rank4color}{RGB}{219, 0, 34}     %
\definecolor{rank5color}{RGB}{205, 0, 47}     %
\definecolor{rank6color}{RGB}{196, 0, 57}     %
\definecolor{rank7color}{RGB}{183, 0, 71}     %
\definecolor{rank8color}{RGB}{170, 0, 83}     %
\definecolor{rank9color}{RGB}{157, 0, 95}     %
\definecolor{rank10color}{RGB}{147, 2, 107}     %
\definecolor{rank11color}{RGB}{131, 1, 118}     
\definecolor{rank12color}{RGB}{121, 0, 131}     
\definecolor{rank13color}{RGB}{107, 0, 144}     
\definecolor{rank14color}{RGB}{99, 0, 155}     
\definecolor{rank15color}{RGB}{85, 0, 167}     
\definecolor{rank16color}{RGB}{71, 0, 181}     
\definecolor{rank17color}{RGB}{62, 3, 193}     
\definecolor{rank18color}{RGB}{51, 0, 204}     
\definecolor{rank19color}{RGB}{37, 0, 216}     
\definecolor{rank20color}{RGB}{25, 0, 229}     
\definecolor{rank21color}{RGB}{6, 0, 242}     
\definecolor{rank22color}{RGB}{0, 0, 255}     

\newcommand{\sq}[1]{\tikz[baseline=\squarebaseline]{\draw[fill=#1, draw=black, line width=\squareborder] (0,0) rectangle (\squaresize,\squaresize);}\hspace{\squarespacing}}
\newcommand{\rankcircle}[1]{\tikz[baseline=(char.base)]{\node[circle, fill=rank#1color, minimum width=\circlesize, minimum height=\circlesize, inner sep=0pt, outer sep=0pt, text=circletextcolor, font=\circlefontsize] (char) {#1};}\hspace{0.2em}}

\definecolor{bailiang}{RGB}{147, 170, 178}        
\definecolor{nextgennn}{RGB}{229, 210, 129}         
\definecolor{honkamj}{RGB}{122, 85, 160}           
\definecolor{loraft}{RGB}{187, 100, 75}          
\definecolor{madeforlife}{RGB}{190, 160, 220}     
\definecolor{lukasf}{RGB}{233, 100, 160}          
\definecolor{lyu}{RGB}{129, 190, 80}                
\definecolor{timh}{RGB}{240, 49, 49}                
\definecolor{vroc}{RGB}{80, 216, 120}              
\definecolor{dutchmasters}{RGB}{220, 80, 229}      
\definecolor{zhuoyuan}{RGB}{255, 120, 0}          
\definecolor{antssyn}{RGB}{220, 220, 220}         
\definecolor{deedsbcv}{RGB}{190, 190, 190}        
\definecolor{fireantsgreedy}{RGB}{160, 160, 160}  
\definecolor{fireantssyn}{RGB}{129, 129, 129}     
\definecolor{synthmorph}{RGB}{100, 100, 100}      
\definecolor{transmorph}{RGB}{0, 0, 255}        
\definecolor{unigradicon}{RGB}{0, 75, 255}      
\definecolor{unigradiconiso}{RGB}{0, 150, 255}  
\definecolor{vfa}{RGB}{0, 220, 255}               
\definecolor{voxelmorph}{RGB}{140, 255, 255}      
\definecolor{zerodisplacement}{RGB}{39, 39, 39}      
\definecolor{resp}{RGB}{0, 0, 0}
\usepackage[disable]{todonotes}
\begin{document}

\title{Encoder-Only Image Registration}

\author{Xiang Chen, Renjiu Hu, Jinwei Zhang, Yuxi Zhang, Xinyao Yu, Min Liu, Yaonan Wang, and Hang Zhang 
\thanks{Manuscript received April 19, 2021; revised August 16, 2021. Copyright © 2021 IEEE. Personal use of this material is permitted. However, permission to use this material for any other purposes must be obtained from the IEEE by sending an email to pubs-permissions@ieee.org.
This work was supported in part by the National Natural Science Foundation of China (grant numbers U22B2050, 62425305, and 62503161), and in part by the Natural Science Foundation of Hunan Province under Grant 2025JJ60389, and in part by the Congressionally Directed Medical Research Programs (CDMRP) under Grant HT9425-25-1-0716. (Corresponding author: Hang Zhang.)}
\thanks{Xiang Chen, Min Liu, Yuxi Zhang, and Yaonan Wang are with the School of Artificial Intelligence and Robotics, Hunan University, Changsha 410082, China. (e-mail: xiangc@hnu.edu.cn, liu\_min@hnu.edu.cn, hnuzyx@hnu.edu.cn, yaonan@hnu.edu.cn). }
\thanks{Renjiu Hu, and Hang Zhang are with Cornell University, New York, USA. (e-mail:rh656@cornell.edu, hz459@cornell.edu).}
\thanks{Jinwei Zhang is with the Department of Electrical and Computer Engineering, Johns Hopkins University, Baltimore, MD 21218, USA (e-mail:jwzhang@jhu.edu). Xinyao Yu is with the Department of Electrical and Computer Engineering, National University of Singapore, Singapore 117583 (e-mail:xinyao.yu@u.nus.edu).}

}

\markboth{Journal of \LaTeX\ Class Files,~Vol.~14, No.~8, August~2021}%
{Shell \MakeLowercase{\textit{et al.}}: Encoder-Only Image Registration}

\IEEEpubid{0000--0000/00\$00.00~\copyright~2021 IEEE}

\maketitle

\begin{abstract}
Learning-based techniques have significantly improved the accuracy and speed of deformable image registration. 
However, challenges such as reducing computational complexity and handling large deformations persist. 
To address these challenges, we analyze how convolutional neural networks (ConvNets) influence registration performance using the Horn-Schunck optical flow equation. 
Supported by prior studies and our empirical experiments, we observe that ConvNets play two key roles in registration: linearizing local intensities and harmonizing global contrast variations.
Guided by these insights, we propose the Encoder-Only Image Registration (EOIR) framework comprising five modifications to existing approaches, to achieve a better accuracy-efficiency trade-off.
EOIR separates feature learning from flow estimation, employing only a 3-layer ConvNet for feature extraction and a set of 3-layer flow estimators to construct a Laplacian feature pyramid, progressively composing diffeomorphic deformations under a large-deformation model.
Results on six datasets across different modalities and anatomical regions demonstrate EOIR’s effectiveness, achieving superior accuracy-efficiency and accuracy-smoothness trade-offs.
With comparable accuracy, EOIR provides better efficiency and smoothness, and vice versa.
The source code of EOIR is available on \href{https://github.com/XiangChen1994/EOIR}{Github}.
\end{abstract}


\section{Introduction}
\IEEEPARstart{I}{mage} registration, which establishes pixel/voxel correspondences between a pair of images and predicts a deformation field for their spatial alignment, is fundamental to medical imaging and computer vision \cite{chen2021deep}. Essential for applications such as medical image segmentation \cite{avants2008symmetric}, motion tracking \cite{fechter2020one}, surgical guidance \cite{su2009augmented}, and diagnostic analysis \cite{chen2013voxel}, precise image registration facilitates accurate disease detection and monitoring, and the progress of therapeutic procedures.

The methodological landscape of image registration is diverse. As categorized in broader reviews of the field \cite{2021sar_review}, techniques span intensity-based methods (e.g., optical flow, mutual information), feature-based methods \cite{xiang2024two,li2025edge,li2025multiscale} (e.g., leveraging points, lines, or deep features), and their combinations. While effective in many domains, these approaches face pronounced challenges in the context of high-resolution volumetric medical images, where computational burden and large deformations are paramount concerns.  

Traditional methods \cite{ashburner2007fast,vercauteren2009diffeomorphic} typically adopt variants of the Horn-Schunck (H-S) style variational formulation, involving a global dense displacement formulation and computationally expensive iterative processes.
Recent advances in convolutional neural networks (ConvNets) \cite{ronneberger2015u} and transformers \cite{dosovitskiyimage,liu2021swin} have enabled learning-based registration methods \cite{balakrishnan2019voxelmorph,hoopes2021hypermorph} to achieve faster performance by amortizing optimization across cohorts, potentially improving accuracy when trained semi-supervised with label supervision.
Despite progress, two major challenges remain for learning-based registration methods, as outlined below.

\begin{figure}[t]
    \centering
    \includegraphics[width=1.\linewidth]{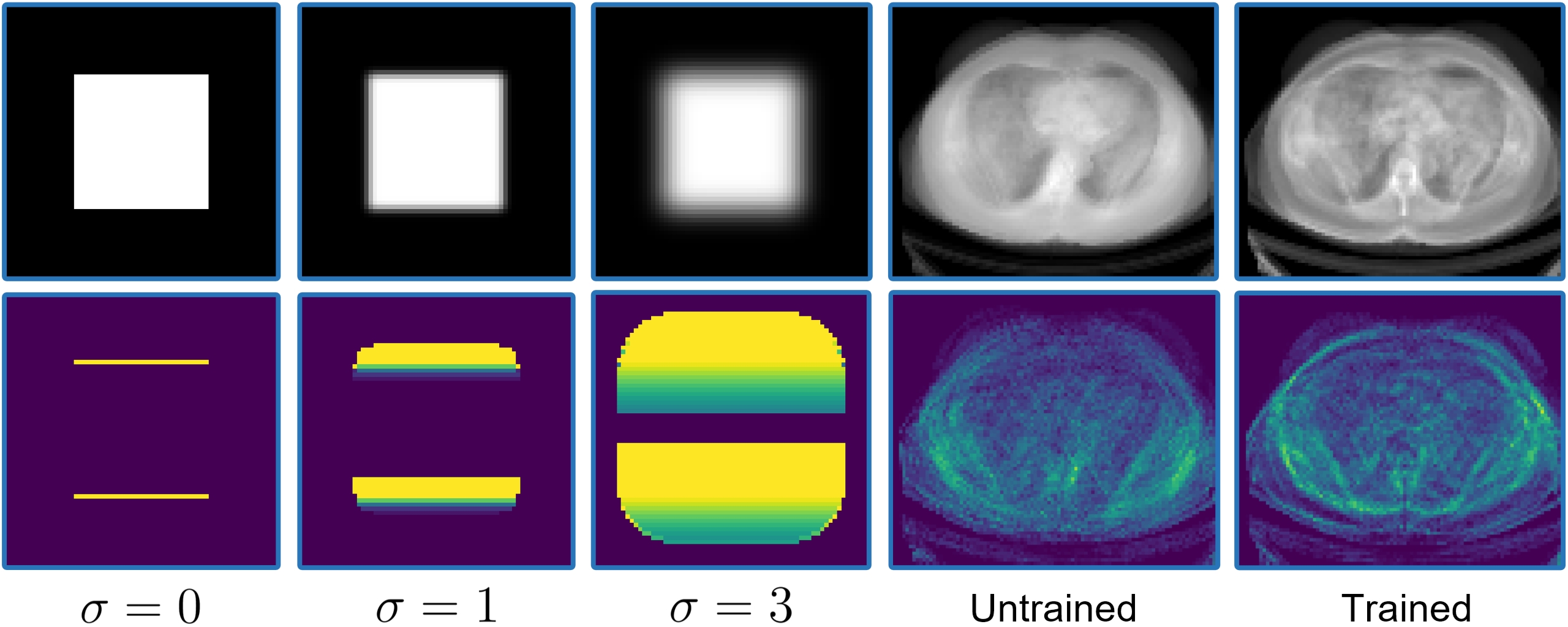}
    \caption{
    Visual demonstration of local intensity linearization.
    The top row shows synthetic and real-world images, while the bottom row presents corresponding heatmaps (values 0 to 1) in the `viridis' color map, where brighter areas indicate better linearization (heatmap generation detailed in the appendix). 
    The first three columns show synthetic examples: a binary square (value 0 and 1) and its Gaussian-blurred versions with $\sigma=1$ and $\sigma=3$.  
    The last two columns display abdominal CT examples, with heatmaps derived from feature maps of untrained and trained ConvNets. 
    Both Gaussian filtering and trained neural networks enhance local intensity linearization.
    }
    \label{fig:conv_vis}
\end{figure}

\IEEEpubidadjcol 
\textbf{Lowering Computational Complexity}:  
Registering volumetric medical images demands considerable computational resources.  
To improve registration accuracy, advanced neural network modules, such as transformers \cite{dosovitskiyimage,liu2021swin} or large convolution kernels \cite{jia2022u,ding2022scaling}, have been proposed, but these modules come with further increased computational demands. 
Despite this, the challenge of reducing computational complexity has been largely overlooked in the literature, with only a few learning-based methods \cite{wang2020deepflash,jia2023fourier,jia2024decoder,yang2024shiftmorph} explicitly addressing the efficiency issue.
Yet, these approaches often lower computational complexity at the cost of accuracy, failing to achieve a practical balance between accuracy and efficiency for volumetric image registration. 

\textbf{Handling Large Deformations}:  
To manage large deformations, many existing approaches employ cascaded network architectures \cite{zhao2019recursive,jia2021learning}, multi-scale coarse-to-fine strategies \cite{mok2021large,meng2024correlation}, or image pyramid structures \cite{zhang2024memwarp,wang_RDP}.  
While effective, these methods often rely on complex architectures, complicating the training process and increasing computational demands.
We argue that the inability to effectively integrate prior knowledge limits the capacity of current neural networks from being fully utilized.

Previous research has shown that incorporating prior knowledge can improve the accuracy-efficiency trade-off in both image segmentation \cite{zhang2023spatially} and image registration \cite{wang2020deepflash,jia2023fourier,chen2024textscf}, which motivates the design of our architecture. 
Building on this principle, we identify two key roles of ConvNets in image registration, based on our empirical observations and prior studies \cite{islam2019much,wang2023robust,zhang2024memwarp}:  
1) \textbf{linearizing local intensities}, ensuring intensity changes vary linearly with spatial coordinates to aid displacement estimation in textureless regions.
2) \textbf{harmonizing global contrast variations}, minimizing intensity discrepancies of the same anatomical regions across different subjects or phases to facilitate alignment.
These two roles help produce image features that satisfy the brightness constancy assumption in the Horn-Schunck (H-S) optical flow equation \cite{horn1981determining} (hereafter referred to as the \textbf{H-S assumption}).

The first role can be empirically observed in Fig. \ref{fig:conv_vis}.  
The left three columns illustrate that simple convolution operations, such as Gaussian filtering, help linearize local intensities of a simple structure (a uniform square here), aiding displacement recovery in textureless regions (the inner areas of the square).
The right two columns demonstrate that trained ConvNet filters achieve superior linearization compared to untrained ones on more complex real-world abdominal data, as evidenced by a wider distribution of linearized locations.
This observation aligns with \cite{islam2019much}, which showed that ConvNets implicitly learn features as a function of spatial coordinates, with deeper layers improving the readout of larger distances.  
Fig. \ref{fig:conv_vis} was created based on the H-S assumption, which relates the intensity change between a pair of images at a given location to the product of the displacement and the local image gradient.  

The second role is supported by \cite{wang2023robust}, which observed that Pearson's correlation between feature maps of different modalities increases with network depth, indicating that deeper ConvNet layers generate features more invariant to input modality.  
Additionally, our prior work \cite{zhang2024memwarp} demonstrated that image features learned for registration tasks can benefit segmentation, creating mutual improvements for both tasks.
Given that ConvNets can bridge differences between modalities and harmonize image intensities into more uniform features across regions (as in segmentation), handling registration tasks with milder contrast variations, such as those addressed in this work, may require fewer convolutional layers and be more computationally efficient.

Based on these observations, we propose the \textbf{Linearization-Harmonization (L-H) assumption} as a design guideline for our registration network: \textit{linearizing local intensities and harmonizing global contrast variations constitute the core roles of ConvNets in deformable image registration.}
The H-S assumption informs us about what features are beneficial for registration, while the L-H assumption explains when a neural network can produce such features.  
To decrease the computational complexity while handling large deformation,  we introduce the following modifications to existing networks under the guidance of H-S and L-H assumption:  
\textbf{M1}: Decoupled feature extraction: unlike traditional learning-based methods \cite{balakrishnan2019voxelmorph,mok2021large,chen2022transmorph} that concatenate moving and fixed images at the network input and process the entire network as a unified flow estimator, we extract feature maps from moving and fixed images independently. This separation of feature extraction from flow estimation enables the generation of image features consistent with the \textbf{H-S} assumption.
\textbf{M2}: Large deformation diffeomorphic framework: since the \textbf{H-S} assumption holds primarily for small displacements, we embed the registration process in a large deformation diffeomorphic framework with a multi-resolution pyramid and scaling-and-squaring integration at each level. The moving image features are progressively warped using the deformation field from the preceding level, so that only residual deformations (ideally smaller than one voxel) are estimated at each stage.
\textbf{M3}: Moving and fixed features are combined using the Hadamard product, which acts as a lightweight cost-volume and naturally conforms to the local matching principle of the \textbf{H-S} assumption.
\textbf{M4}: Multi-level similarity loss with Gaussian smoothing: we compute image similarity losses across multiple resolution levels, applying Gaussian smoothing before down-sampling to improve gradient behavior in homogeneous regions (corresponding to both \textbf{H-S} and \textbf{L-H} assumption).
\textbf{M5}: Lightweight encoder with shallow convolutional blocks: based on the above design, the network extracts sufficiently discriminative features using only a few convolutional blocks for mono-modal image registration and simple multi-modal image registration. 


To accommodate these modifications, we propose the \textbf{E}ncoder-\textbf{O}nly \textbf{I}mage \textbf{R}egistration (\textbf{EOIR}) framework to address the aforementioned challenges (see Fig. \ref{fig:eoir_framework} for an overview of the EOIR architecture). The name EOIR reflects the framework's simplicity, as it utilizes only an encoder, foregoing a more complex encoder-decoder structure. While such terms may vary across contexts, we emphasize this minimal design to highlight its efficiency.
The major contributions of this work are summarized as follows:  
\textcolor{resp}{
\begin{itemize}  
    \item We propose EOIR, a registration framework whose architecture is explicitly derived from the \textbf{H-S} and \textbf{L-H} assumptions. This is realized through five key design choices, including decoupled feature extraction, a diffeomorphic pyramid with warped feature propagation, Hadamard-based feature matching, and multi-level loss with Gaussian smoothing.
    \item The above design leads to an exceptionally efficient model. By decoupling feature extraction and using lightweight matching, EOIR achieves state-of-the-art efficiency-accuracy trade-offs, enabling higher accuracy at comparable complexity or drastically lower complexity without sacrificing accuracy.
    \item Evaluated on six diverse datasets, EOIR demonstrates strong generalization. It secured 2nd place in the Learn2Reg LUMIR Challenge 2024 \cite{chen2025lumir} with a sub-1MB model trained in two days on ~4,000 subjects, proving its readiness for large-scale, accurate registration.
\end{itemize} }\todo{\scriptsize R3.1}

The remainder of this paper is organized as follows. Section II reviews related work. Section III details our proposed EOIR framework. Section IV describes the experimental setup and presents a comprehensive analysis of the results. Section V concludes the paper.

\section{Related Work}
The proposed EOIR framework falls under the category of learning-based registration models.  
In this section, we first review classic learning-based registration methods, followed by computationally efficient registration techniques, and conclude with models designed to handle large deformations.

\subsection{Learning-Based Image Registration}
\label{sec:related_1}
Deformable image registration (DIR) is traditionally formulated as an energy optimization problem where dissimilarity between moving $\mathbf{I}_m$ and fixed $\mathbf{I}_f$ images is quantified using a dissimilarity function $s(\cdot)$. 
To counteract the ill-posed nature of DIR, a regularization term $r(\cdot)$ constrains the deformation field. 
\textcolor{resp}{While traditional methods directly optimize the deformation field through gradient descent \cite{avants2008symmetric} or discrete optimization \cite{heinrich2013mrf,zhang2025voxelopt}, learning-based approaches \cite{balakrishnan2019voxelmorph,hu2022cross,zheng2025plug,han2023diffeomorphic,lyu2018super,sun2024medical,zhang2022learning,zhang2024heteroscedastic} optimize the expected loss function to derive neural network weights $\theta$ from a collection of image pairs $D$}\todo{\scriptsize R6.3}, as formulated below:
\begin{equation}
\hat{\theta} = \argmin_{\theta} \{ \mathbb{E}_{(\mathbf{I}_f,\mathbf{I}_m) \sim D} [\mathcal{L}(\mathbf{I}_f, \mathbf{I}_m, g_{\theta}(\mathbf{I}_f,\mathbf{I}_m))] \}.
\label{eq:nn_optimization}
\end{equation}
In this equation, $\mathcal{L}(\mathbf{I}_f, \mathbf{I}_m, \mathbf{u}) = s(\mathbf{I}_f, \mathbf{I}_m \circ \phi) + r(\mathbf{u})$ denotes the loss function. 
Here, $\mathbf{I}_m \circ \phi$ represents the warping of the moving image by the deformation field $\phi = \mathbf{I}_d + \mathbf{u}$, where $\mathbf{I}_d$ is the identity transformation grid. 

Using Eq.~\eqref{eq:nn_optimization} for unsupervised learning is much faster than traditional energy optimization methods and may benefit from label supervision such as segmentation loss, which may further increase the accuracy of anatomical alignment. 
VoxelMorph \cite{balakrishnan2019voxelmorph}, a pioneering learning-based model, entangles feature extraction and flow estimation in a single U-Net architecture \cite{ronneberger2015u}, processing volumetric brain MR images in seconds. 
Following VoxelMorph, \cite{dalca2019unsupervised} introduces scaling and squaring layers \cite{ashburner2007fast} to ensure diffeomorphism. 
Vision transformers \cite{liu2021swin} have also been incorporated into frameworks like TransMorph \cite{chen2022transmorph} and H-ViT \cite{ghahremani2024h}.
Symmetric registration networks \cite{mok2020fast}, multi-channel architectures \cite{chen2021deepDiscontinuity}, dual-stream networks \cite{kang2022dual}, large-kernel convolutions \cite{jia2022u,chen2024textscf}, and cascaded networks \cite{zhao2019recursive,mok2021large} further contribute to progress in the field.
However, these improvements often lead to an exponential increase in parameters, raising computational demands, which can be challenging in resource-constrained clinical settings such as limited GPU memory or large volumetric datasets \cite{liu2024finite}.

\begin{figure}[ht]
    \centering
    \includegraphics[width=0.95\linewidth]{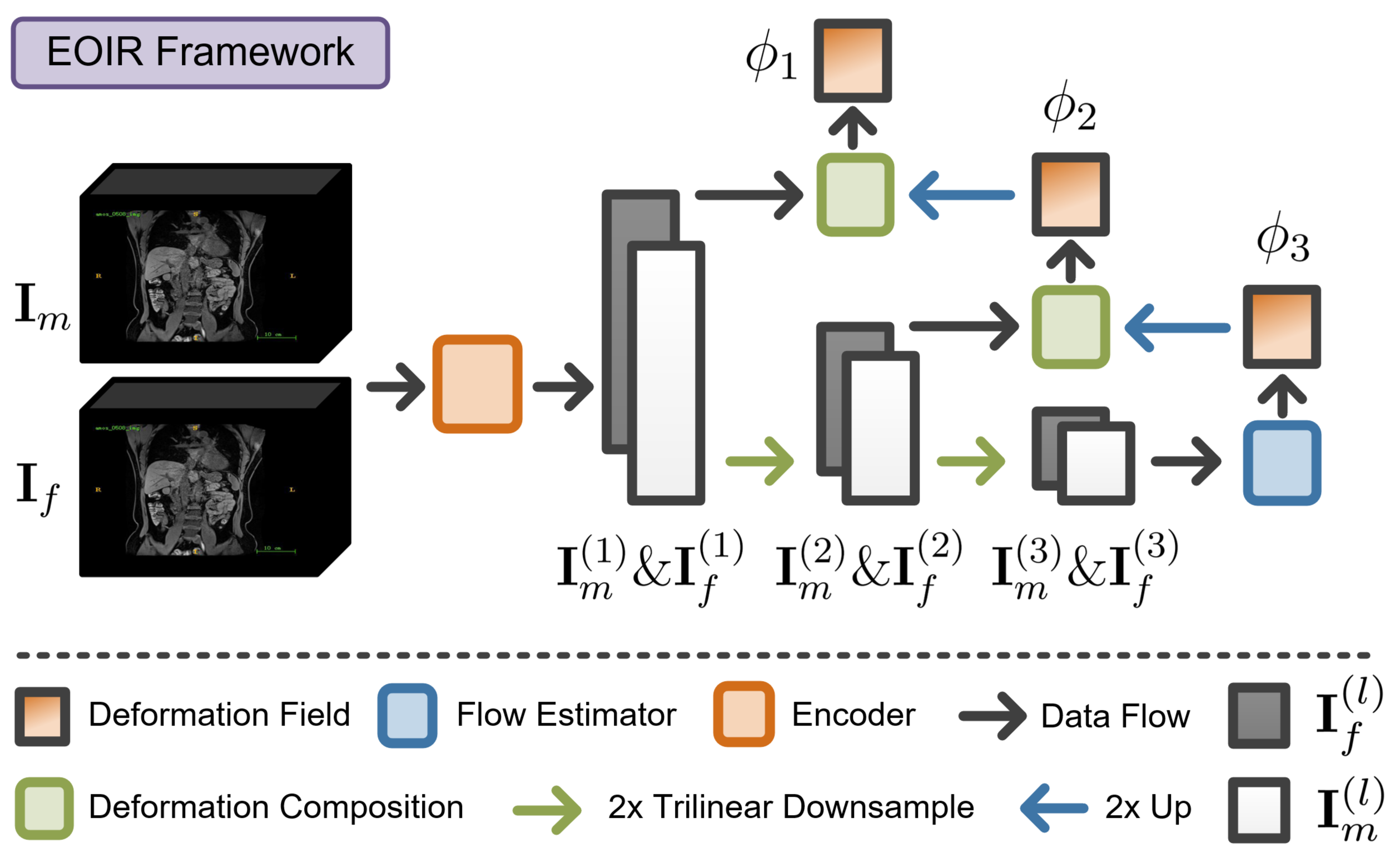}
    \caption{  
    Architecture of the EOIR framework. The three-level pyramid operates as follows: (1) Features $I_m^{(l)}$ and $F_m^{(l)}$ are independently extracted (via encoder) and downsampled. (2) Deformation fields $\phi_1$–$\phi_3$ are estimated per level via flow estimators. (3) Deformations are composed across levels. This process breaks large deformations into a sequence of small, H–S-compliant residual steps, enabling robust registration. See §\ref{sec:eoir_arch} for details.
    }  
    \label{fig:eoir_framework}
    \vspace{-2ex}
\end{figure}

\subsection{Computationally-Efficient Image Registration}
\label{sec:related_2}

Pursuing computational efficiency in image registration has been approached through simplified representations in traditional methods and, more recently, through specialized network architectures in deep learning. Exemplifying the former, Albu \cite{albu2016low} transformed the 2D problem into 1D signal alignment via integral projections. Meanwhile, the explicit design of inherently efficient deep networks remains less explored, with only a few works directly addressing this goal \cite{wang2020deepflash,jia2023fourier,jia2024decoder,yang2024shiftmorph,zhang2025voxelopt}. Some efficiency gains are achieved as a byproduct of specific architectural choices \cite{balakrishnan2019voxelmorph,chen2024textscf,zhang2024slicer}.
These pioneering efforts inspired EOIR's design, demonstrating how integrating prior knowledge can reduce complexity while maintaining accuracy, or even improve it.  

DeepFlash \cite{wang2020deepflash} approximates the original displacement space using a low-dimensional, band-limited space, performing neural network inference within this constrained domain.  
This accelerates training and inference without sacrificing accuracy compared to VoxelMorph, based on the assumption that flow fields inherently lack high frequencies in the Fourier domain.  
Similarly, FourierNet \cite{jia2023fourier} builds on this prior but improves efficiency by employing a model-driven decoder to better leverage the band-limited approximation, achieving a superior accuracy-efficiency trade-off.  
ShiftMorph \cite{yang2024shiftmorph} skips the Fourier transform but applies a similar concept, operating at lower spatial resolutions for greater efficiency.  
LessNet \cite{jia2024decoder} eliminates the encoder entirely, significantly reducing network parameters while maintaining accuracy comparable to VoxelMorph \cite{balakrishnan2019voxelmorph} and TransMorph \cite{chen2022transmorph}.  

Additionally, some models \cite{chen2024textscf,chen2025ideal,zhang2024slicer,he2025samir} incorporate other forms of prior knowledge to further improve efficiency.  
TextSCF \cite{chen2024textscf} uses a large visual-language model to enhance inter-regional anatomical understanding, outperforming models without priors in both efficiency and accuracy \cite{balakrishnan2019voxelmorph, chen2022transmorph, mok2021large, jia2022u}.  
The Slicer Network \cite{zhang2024slicer}, though designed for general medical image analysis, employs edge-preserving adaptive filters to expand the effective receptive field \cite{luo2016understanding}, leading to better accuracy-efficiency trade-offs.  
Different from the above-mentioned approaches, our framework, EOIR, achieves a more favorable accuracy-efficiency balance by novelly integrating the H-S and L-H assumptions, strategically disentangling feature extraction from flow estimation, and employing a lightweight convolutional encoder.

\subsection{Large-Deformation Image Registration}  
Most learning-based models discussed in \S\ref{sec:related_1} and \S\ref{sec:related_2} exhibit limited performance on datasets involving large deformations.  
Even methods employing vision transformers or large convolutional kernels to expand the effective receptive field often fail to address such deformations adequately.  
To overcome this limitation, two primary architectural strategies have emerged: (1) recursive or cascaded flow estimators \cite{zhao2019unsupervised,jia2021learning,zhang2025unsupervised,zhang2025unsupervised}; (2) coarse-to-fine image pyramids \cite{mok2021large,zhang2024memwarp,meng2024correlation}.  

The first approach, exemplified by VTN \cite{zhao2019unsupervised} and VR-Net \cite{jia2021learning}, employs cascaded sub-networks that iteratively refine deformations through sequential steps, with each step generating a full-resolution deformation field.
The second strategy, adopted in LapIRN \cite{mok2020fast}, IIRP-Net \cite{ma2024iirp}, MemWarp \cite{zhang2024memwarp}, and RDP \cite{wang_RDP} progressively refines deformation fields using a coarse-to-fine pyramid framework, which often achieves better efficiency and large-deformation handling.
CorrMLP \cite{meng2024correlation} further enhances this paradigm by integrating a correlation-aware multi-window MLP block into its coarse-to-fine architecture.

However, these methods typically lack explicit prior knowledge, requiring additional computational resources to process cascaded sub-networks or multi-scale features.
In contrast, our EOIR framework is designed with a more efficient feature pyramid, guided by the integrated H-S and L-H assumptions as architectural priors, which enables a superior balance between accuracy and efficiency.

\section{Methodology}  
\label{sec:method}  
In this section, we begin with preliminaries which review the H-S optical flow equation and explain how neural networks operating under the L-H assumption facilitate the production of image features that satisfy the H-S assumption.  
We then introduce the large-deformation diffeomorphic model, followed by a detailed description of our EOIR framework, including its network architecture and loss functions.  

\begin{figure}[t]
    \centering
    \includegraphics[width=0.9\linewidth]{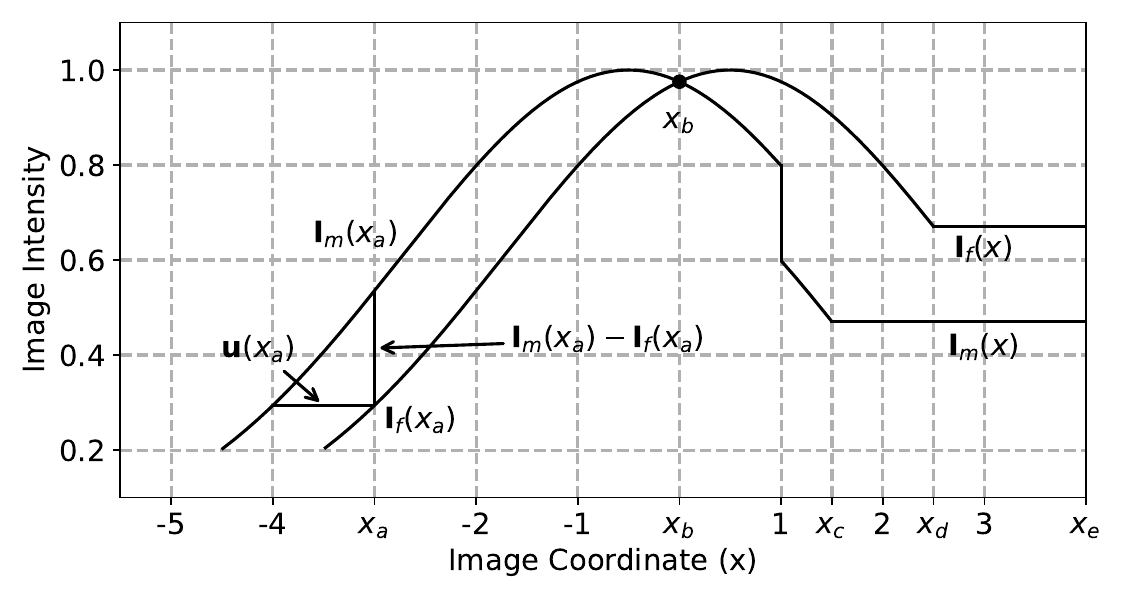}
    \vspace{-2ex} 
    \caption{
    A one-dimensional analogue of Eq.~\eqref{eq:horn_schunck} holds at \(x_a\), where \(\frac{\mathbf{I}_m(x_a) - \mathbf{I}_f(x_a)}{\mathbf{u}(x_a)} \approx \frac{d\mathbf{I}_f}{dx}(x_a)\).  
    Here, \(\mathbf{I}_m(x)\) is generated by translating \(\mathbf{I}_f(x)\) horizontally by \(-1\) and adding a global bias of \(-0.2\) for \(x > 1\).  
    However, the displacement cannot be determined at \(x_b\) (where \(\mathbf{I}_m(x_b) - \mathbf{I}_f(x_b) \approx 0\)), between \(x=2\) and \(x_d\) (where a global bias is applied), or between \(x_d\) and \(x_e\) (where \(\|\nabla \mathbf{I}_f\| = \|\nabla \mathbf{I}_m\| = 0\)).  
    Additional constraints are required to propagate displacements from surrounding regions.  
    }
    \label{fig:horn_schunck}
\end{figure}

\subsection{Preliminary}
\subsubsection{Horn–Schunck (H-S) Equation}
\label{sec:method_hs_eq}
The original H-S equation \cite{horn1981determining} relates the displacement field \(\mathbf{u}(p)\) between images \(\mathbf{I}_m\) and \(\mathbf{I}_f\) to their intensity difference and gradient under sufficient spatial sampling:
\begin{equation}
    \nabla \mathbf{I}_f(p) \cdot \mathbf{u}(p) = \mathbf{I}_m(p) - \mathbf{I}_f(p),
    \label{eq:horn_schunck}
\end{equation}
where \(\nabla \mathbf{I}_f(p) = \left[ \frac{\partial \mathbf{I}_f}{\partial x}, \frac{\partial \mathbf{I}_f}{\partial y}, \frac{\partial \mathbf{I}_f}{\partial z} \right]_p\) is the spatial gradient at voxel \(p \in \Omega\). 
Equation~\eqref{eq:horn_schunck} holds when \(\|\mathbf{u}(p)\|^2 < 1\) voxel, achievable through sufficient spatial downsampling (see \(x_a\) in Fig.~\ref{fig:horn_schunck}).  
To satisfy this small-displacement requirement, we adopt a Laplacian feature pyramid \cite{mok2021large,zhang2024memwarp}, where the number of pyramid levels \(n\) is determined by the maximum displacement \(d_{\text{max}}\) in the dataset:  
\begin{equation}  
n > \log_2(d_{\text{max}}) + 1,  
\label{eq:n_level_compute}
\end{equation}  
ensuring deformations up to \(d_{\text{max}}\) are resolved.
The number of pyramid levels is determined by the maximum displacement in a given dataset.  
While adjusting the levels can reduce complexity, we adopt a 5-level pyramid for simplicity, which generalizes well across most datasets. (See \S\ref{sec:results_effects_pyramid_n} for a comparison of the effects of varying pyramid levels across datasets.)

Despite the use of a pyramid, Eq.~\eqref{eq:horn_schunck} may fail even with sub-voxel displacements in three scenarios:
1) at locations with insufficiently large image gradients, indicating flat or noisy regions where no local constraints can be imposed ($x_d$ to $x_e$ in Fig. \ref{fig:horn_schunck});
2) when the intensity difference between the moving and fixed images at certain locations is close to zero ($x_b$ in Fig. \ref{fig:horn_schunck});
3) when a global intensity bias is added to the moving or fixed image, as between $x = 2$ and $x_d$ in Fig. \ref{fig:horn_schunck}).
Voxels in these scenarios can be addressed by feature linearization via a neural network and global smoothness constraints through a loss function. 
The former offers features with local intensity linearization, while the latter can assist in propagating displacements from surrounding valid locations.

\subsubsection{Large-Deformation Diffeomorphic Model}

For large diffeomorphic deformations, a widely used approach \cite{arsigny2006log,ashburner2007fast} involves utilizing a stationary velocity field (SVF) $\mathbf{v}$. 
This field is integrated over unit time from $t = 0$ to $t = 1$ to produce the final deformation field $\phi^{(1)}$, starting from the identity transformation $\phi^{(0)} = \mathbf{I}_d$. 
The integration is governed by the ordinary differential equation (ODE):
\begin{equation}
    \frac{\mathrm{d} \phi^{(t)}}{\mathrm{d} t} = \mathbf{v}(\phi^{(t)}).
    \label{eq:ldd_one}
\end{equation}

However, as noted in Section \S\ref{sec:method_hs_eq}, modeling displacements up to a maximum \(d_{max}\) requires a Laplacian feature pyramid of \(n\) levels to ensure \(2^{n-1} > d_{max}\) (Eq.~\ref{eq:n_level_compute}). 
Given that \(n\) residual deformation fields are estimated at each pyramid level and must be composed to form the final deformation field, a single SVF is insufficient to adequately model the dynamics across different pyramid levels. 
Therefore, we utilize \(n\) SVFs, each corresponding to a different pyramid level, to govern the evolution of the deformation. 
Consequently, with \(\mathbf{v}^{(t)}\) as a piece-wise constant function representing the velocity field at time \(t\), the deformation evolves as the following ODE:
\begin{equation}
    \frac{\mathrm{d} \phi^{(t)}}{\mathrm{d} t} = \mathbf{v}^{(t)}(\phi^{(t)}).
    \label{eq:ldd_n}
\end{equation}
The integration of Eq.~\eqref{eq:ldd_n} can be discretized into a finite number of steps, expressed as:
\begin{equation}
    \phi^{(t+\Delta t)} = (\mathbf{I}_d + \Delta t \mathbf{v}^{(t)}) \circ \phi^{(t)},
\end{equation}
where \(\Delta t\) is the integration step size. 
This method models large deformations as a sequence of small deformations at each step. 
To ensure accurate approximation, the step number must be large enough to capture each deformation sufficiently. 
    
\subsubsection{Integration and Deformation Field Composition}
In learning-based registration models such as VoxelMorph \cite{balakrishnan2019voxelmorph}, which output a single deformation field for an image pair, Eq.~\eqref{eq:ldd_one} can parametrize the deformation field.
The \textit{scaling and squaring} method, derived from Lie Theory, is an efficient integration solution. 
It's widely used for rapid integration in learning-based registration models \cite{dalca2019unsupervised,mok2021large}, enabling efficient computation of diffeomorphic deformations.

Models using a feature pyramid, which must compose multiple deformation fields from coarse to fine, require more careful handling of integration and composition.
For example, LapIRN \cite{mok2021large} and MemWarp \cite{zhang2024memwarp} use a Laplacian feature pyramid and compose deformation fields across levels by addition.
While computationally less demanding, this approach can result in a poorer trade-off between deformation smoothness and registration accuracy \cite{chen2024textscf}, as well as slower convergence and a tendency to settle at sub-optimal local minima \cite{vercauteren2009diffeomorphic}.

To address this issue, we propose discretizing the unit time into a substantial number of smaller intervals, specifically $n\times m$ steps, where $n$ is the number of pyramid levels, and $m$ is the number of steps within each level corresponding to the pyramid hierarchy.
Let $\phi^{t_{ij}}$ denote the $j^{th}$ step of the $i^{th}$ level, the final deformation field $\phi$ can be approximated as:
\begin{align}
    \phi \approx & ~ \phi^{t_{11}} \circ \phi^{t_{12}} \circ \ldots \circ \phi^{t_{1m}} \circ \nonumber \\ 
            & ~ \phi^{t_{21}} \circ \phi^{t_{22}} \circ \ldots \circ \phi^{t_{2m}} \circ \nonumber \\ 
            & ~ \ldots \nonumber \\ 
            & ~ \phi^{t_{n1}} \circ \phi^{t_{n2}} \circ \ldots \circ \phi^{t_{nm}}.
    \label{eq:xia_ji_ba_xie}
\end{align}
While naively computing Eq.~\eqref{eq:xia_ji_ba_xie} demands $\mathcal{O}(nm)$ complexity and may prove computationally burdensome, employing the \textit{scaling and squaring} method by \cite{arsigny2006log} for each level significantly reduces this to $\mathcal{O}(n\log m)$.

\subsection{Network Architecture}
\label{sec:eoir_arch}

The design of the EOIR architecture is guided by the H-S and L-H assumptions. 
It comprises three main components: an encoder for feature extraction, a set of flow estimators, and a deformation field composition module. In the following sections, we detail each component and conclude with a discussion of the loss function for deep supervision \cite{lee2015deeply}.

\subsubsection{Encoder}
Different from previous research \cite{balakrishnan2019voxelmorph,chen2022transmorph}, EOIR disentangle the feature extraction and flow estimation steps, with a light-weight encoder to extract features from the moving and fixed images separately (M1).
Based on the H-S assumption, we construct a large deformation diffeomorphic framework (Fig. \ref{fig:eoir_framework}, M2).  
To model large-deformation diffeomorphic transformations, we approximate the final deformation field via integration over small intervals using Eq.~\eqref{eq:xia_ji_ba_xie}, where \(n\) pyramid levels each contain \(m\) intervals.  
Each interval within a pyramid level follows the small-deformation model \(\phi(p) = p + \mathbf{u}(p)\).  
Despite these design choices, Fig. \ref{fig:horn_schunck} illustrates scenarios where displacement determination remains ambiguous, even under sufficient downsampling (with \(\|\mathbf{u}(p)\|^2 < 1\) voxel), due to vanishing gradients or contrast variations.  

The proposed L-H assumption, grounded in empirical observations from both synthetic and clinical data (Fig. \ref{fig:conv_vis}) and prior research \cite{islam2020much,wang2023robust,zhang2024memwarp}, explains that trained ConvNets for image feature extraction can effectively linearize and harmonize local intensities in mono-modal registration tasks.  
Additionally, Fig. \ref{fig:conv_vis} demonstrates a considerable increase in the number and distribution of valid locations satisfying the H-S assumption when using trained ConvNets.  

\begin{figure}[htbp]
    \centering
    \includegraphics[width=1.0\linewidth]{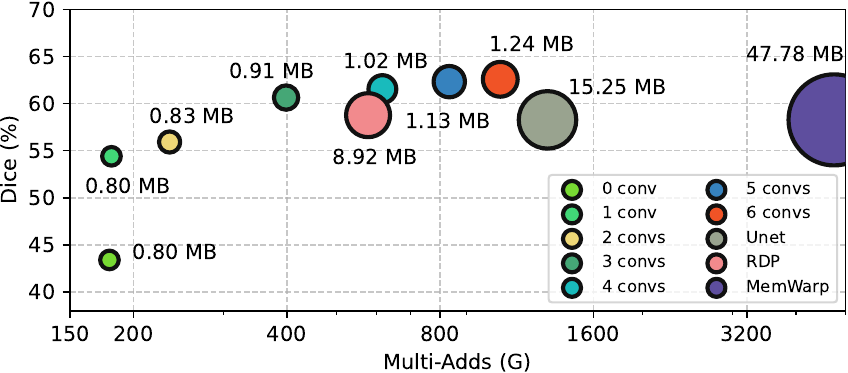}
    \caption{
    Visual comparison of the trade-off between avg. Dice and computational complexity for varying numbers of conv layers in the EOIR encoder ($n_c$ from 0 to 6), alongside top-performing pyramid methods RDP \cite{wang_RDP} and MemWarp \cite{zhang2024memwarp} on the abdomen dataset. 
    Circle size and labels indicate network parameter size, and multi-adds (G) are plotted on a logarithmic x-axis. (see appendix for further metric details). This comparison highlights the effects of our \textbf{M5}.
    }
    \label{fig:nconvs_dsc}
    \vspace{-2ex}
\end{figure}

Guided by H-S and L-H assumptions and the small-deformation model for each interval, we propose that local intensity linearization within small voxel neighborhoods is sufficient to recover large diffeomorphic deformations.  
Thus, we employ only three learnable convolutional layers (M5), rather than a complex encoder-decoder, to efficiently act as local intensity linearizers and contrast harmonizers, achieving an optimal accuracy-efficiency trade-off.  
While additional layers (\(n_c > 3\)) could improve registration accuracy, the marginal gains diminish rapidly beyond three layers (see Fig. \ref{fig:nconvs_dsc}).

\begin{figure*}[htbp]
    \centering
    \includegraphics[width=0.95\linewidth]{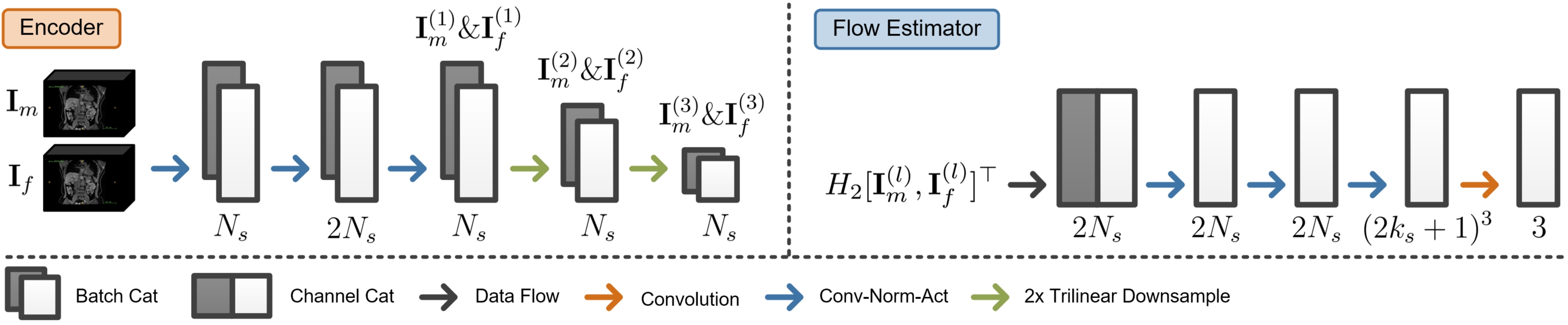}
    \caption{ 
        Visual illustration of the components of the encoder and flow estimator in EOIR. 
        To illustrate the encoder structure, we use a three-level feature pyramid, which consists of three Conv-Norm-Act blocks and two trilinear downsampling layers, producing three pairs of moving and fixed images at different scales. 
        Each pyramid level's flow estimator shares the same structure but with different weights; it consists of a Hadamard transformation, three Conv-Norm-Act blocks, and a single convolution to produce a residual displacement field at that level. In our experiments, we empirically set $K_s=1$.
    }
    \label{fig:encoder_flow_estimator}
\end{figure*}

The encoder comprises three Conv-Norm-Act blocks, each containing a learnable $3\!\times\!3\!\times\!3$ convolutional layer, instance normalization \cite{ulyanov2016instance}, and a ReLU activation.  
As shown in Fig.~\ref{fig:encoder_flow_estimator} (left), the encoder begins with $N_s$ channels, expands to $2N_s$ via an inverted bottleneck design \cite{sandler2018mobilenetv2}, and contracts back to $N_s$.  
Combined with  $n-1$ rounds of $2\times$ spatial downsampling, this design ensures effective local intensity linearization and contrast harmonization.

\subsubsection{Flow Estimator}

With the locally linearized and harmonized feature maps from the encoder, we apply $2\times$ trilinear downsampling to the feature maps $n-1$ times, resulting in an $n$-level feature pyramid. 
This pyramid is structured to handle displacements with a maximum of $d_{max} < 2^{n-1}$ (Eq.~\ref{eq:n_level_compute}). 
At each pyramid level, a flow estimator, consisting of a Hadamard transform layer \cite{pratt1969hadamard} and several conv blocks, generates the residual flow. 
The flow estimator's structure is depicted in the right panel of Fig. \ref{fig:encoder_flow_estimator} and further detailed below.

The Eq.~\eqref{eq:horn_schunck} shows that displacements are linked to the intensity differences between moving and fixed images. 
Rather than using a network to learn this difference, we explicitly apply a Hadamard transform to the feature maps (M3):
\begin{equation}
    H_2[\mathbf{I}_m^{(l)}, \mathbf{I}_f^{(l)}]^{\top} = [\mathbf{I}_m^{(l)} + \mathbf{I}_f^{(l)}, \mathbf{I}_m^{(l)} - \mathbf{I}_f^{(l)}]^{\top},
    \label{eq:deformation_composition}
\end{equation}
where $\mathbf{I}_m^{(l)}$ and $\mathbf{I}_f^{(l)}$ are the moving and fixed image feature maps at pyramid level $l$, and $H_2 = \begin{bmatrix} 1 & 1 \\ 1 & -1 \end{bmatrix}$. 
After the transform, the moving and fixed features are stacked along the channel dimension, processed with three Conv-Norm-Act blocks, and a linear layer on each voxel produces final displacement fields.

\subsubsection{Deformation Field Composition}
\label{sec:def_comp}
To build the large deformation diffeomorphic framework (M2), deformation fields composition step is essential to build the pyramid. We represent the velocity field at time \( t \) by a piece-wise constant function, \( \mathbf{v}^{(t)} \), where the number of pieces corresponds to the number of feature pyramid levels $n$. 
This yields multiple stationary ODEs (Eq. \eqref{eq:ldd_n}) across different periods within the unit time. 
Additionally, under the H-S assumption that the displacement difference between adjacent periods is minimal, we can employ Eq.~\eqref{eq:deformation_composition} to approximate the deformation composition process throughout our feature pyramid, thereby promoting a diffeomorphic transformation.

The deformation field composition at the \(l^{th}\) level of the pyramid involves the following steps. 
Starting with feature maps \( \mathbf{I}_m^{(l)} \) and \( \mathbf{I}_f^{(l)} \), and the deformation field \( \phi_{l+1} \) from the previous level:
\begin{align}
    \tilde{\phi}_{l+1} & = \text{up}(\phi_{l+1}),  \\
    \mathbf{u}_l & = f_{e}^{(l)}(\mathbf{I}_m^{(l)} \circ \tilde{\phi}_{l+1},\mathbf{I}_f^{(l)}),  \\
    \Delta \phi_{l} & = \text{exp}(\mathbf{u}_l),  \\
    \phi_{l} & = \tilde{\phi}_{l+1} \circ \Delta \phi_{l},
\end{align}
where \(\text{up}(\cdot)\) denotes \(2\times\) trilinear upsampling and scaling, \(\text{exp}(\cdot)\) refers to the scaling and squaring function applied to the displacement field, and \(\circ\) denotes the warping function. 
The \(\mathbf{u}_l\) represents the residual displacement field at level \(l\), and $f_e^{(l)}(\cdot,\cdot)$ is the flow estimator at level \(l\).


\subsubsection{Overall Framework \& Deep Supervision}
\label{sec:deep_sup}

With the encoder, flow estimator, and deformation field composition components in place, we have constructed the EOIR framework. 
This framework is visually illustrated in Fig. \ref{fig:eoir_framework} using a 3-level pyramid.
We optimize the network using a multi-scale loss function applied across the registration pyramid. At each level, the total loss comprises a similarity term, which quantifies the dissimilarity between the warped moving image and the fixed image, and a regularization term, which enforces smoothness in the deformation fields, following the convention of \cite{balakrishnan2019voxelmorph, chen2024textscf}.
The total loss is calculated as an exponentially decayed weighted sum across all levels:
\begin{equation}
    \mathcal{L} = \sum_{l=1}^{n}\frac{1}{2^{l-1}}\left[s\left(d(\mathbf{I}_m,l) \circ \phi_{l}, d(\mathbf{I}_f,l)\right) + \lambda r(\mathbf{u}_l)\right].
    \label{eq:loss}
\end{equation}
Here, $n$ represents the number of pyramid levels,  $s(\cdot, \cdot)$ denotes the dissimilarity function, $d(\cdot, l)$ downsamples the input image by a factor of $2^{l-1}$ (Gaussian smoothing is applied before downsampling, M4), and $r(\cdot)$ represents the smoothness regularization function applied to the displacement field $\mathbf{u}_l$ before scaling and squaring.
We use $r(\mathbf{u}_l) = \|\nabla \mathbf{u}_l\|^2$ as the regularization function, with $\lambda$ as its coefficient.

\section{Experiments \& Results}

In this section, we evaluate the proposed EOIR against state-of-the-art image registration methods across six datasets, covering various imaging modalities, input constraints, and anatomies.
The following subsections detail the datasets, implementation details, baseline methods, and evaluation metrics.
We then present qualitative and quantitative results, including analyses of accuracy-efficiency trade-offs, accuracy-smoothness comparisons, and an ablation study.
The source code of EOIR is available on \href{https://github.com/XiangChen1994/EOIR}{Github}. 

\subsection{Datasets}
The datasets span both computed tomography (CT) and magnetic resonance imaging (MRI) modalities, with inter-subject and intra-subject settings, as well as unsupervised and semi-supervised configurations that include segmentation masks.
In summary, we use the semi-supervised Abdomen CT dataset for inter-subject registration \cite{xu2016evaluation}, the semi-supervised OASIS dataset \cite{marcus2007open}, the unsupervised large-scale LUMIR dataset \cite{dufumier2022openbhb,taha2023magnetic,marcus2007open} for inter-subject brain MR image registration, the ACDC dataset \cite{bernard2018deep} for cardiac image registration, the HippocampusMR \cite{simpson2019large} for Hippocampus MR image registration, and the RGB-IR dataset \cite{tang2022piafusion} for multi-modality 2D natural image registration.

\subsubsection{Abdomen CT Dataset}
We employ an abdominal CT dataset comprising 30 scans, each annotated with segmentation masks for 13 anatomical structures \cite{xu2016evaluation}. The dataset is split into 20 training, 3 validation, and 7 test scans, yielding 380 training, 6 validation, and 42 test pairs respectively. All images are resampled to 2 mm isotropic resolution and standardized to a size of $192 \times 160 \times 256$.

\subsubsection{ACDC Dataset} 
We evaluate our method on the ACDC cardiac MR dataset \cite{bernard2018deep}, which includes 80 training, 20 validation, and 50 test subjects. Each subject provides end-diastole (ED) and end-systole (ES) images with ground-truth segmentations of the left ventricle blood pool, myocardium, and right ventricle. Registration is performed in both ED-to-ES and ES-to-ED directions, resulting in 160 training, 40 validation, and 100 test pairs. All images are preprocessed to $128 \times 128 \times 16$ with a voxel spacing of $1.8 \times 1.8 \times 10$ mm³.

\subsubsection{OASIS Dataset}
For semi-supervised inter-subject brain MR registration, we use the OASIS dataset from Task 3 of the Learn2Reg 2021 challenge \cite{marcus2007open, hering2022learn2reg}. It contains 414 T1-weighted brain MRI scans, of which 394 unpaired scans are used for training. Validation and leaderboard ranking employ 19 paired images generated from 20 scans\footnote{\url{https://learn2reg.grand-challenge.org/evaluation/task-3-validation/leaderboard/}}. All images are preprocessed with bias correction, skull stripping, alignment, and cropping to $160 \times 192 \times 224$.

\subsubsection{LUMIR Dataset}
The LUMIR dataset \cite{dufumier2022openbhb, taha2023magnetic, marcus2007open} is designed for large-scale unsupervised brain MR registration as part of Learn2Reg 2024 Task 2. It includes 3,384 training subjects and 40 validation subjects, with 10 training subjects manually annotated with anatomical landmarks to generate 38 validation pairs. All images are provided in NIfTI format, resampled, cropped to the region of interest, and standardized to $160 \times 224 \times 192$ with $1 \times 1 \times 1$ mm³ spacing.

\subsubsection{HippocampusMR Dataset}  
The HippocampusMR dataset \cite{simpson2019large} focuses on inter-subject hippocampus MR registration and was part of the Learn2Reg 2020 challenge. We split the data into 200 training, 20 validation, and 40 test subjects. All images are preprocessed to $64 \times 64 \times 64$ with isotropic $1 \times 1 \times 1$ mm³ spacing.

\subsubsection{RGB-IR Dataset}
The RGB-IR dataset \cite{tang2022piafusion} contains 1,354 paired 2D RGB and infrared images. Following \cite{zheng2025plug}, random affine and deformable transformations are applied to generate registration pairs. The dataset is divided into 1,274 training, 30 validation, and 50 test pairs, with all images cropped to $256 \times 256$.

\subsection{Implementation Details and Baseline Methods}
\subsubsection{Training Details}
In this study, all models were developed using the PyTorch library in Python, executed on a system with an A100 GPU. 
The Adam optimizer was employed for training, with an initial learning rate of $1e-4$ and a polynomial learning rate scheduler with a decay rate of 0.9. 
The channel parameters for EOIR's encoder and flow estimator were set to $N_s=32$ and $k_s=1$, unless otherwise specified. 
The number of image pyramid levels is set to $n=5$, and the number of time steps in each period for Eq.~\eqref{eq:xia_ji_ba_xie} is set to $m=7$, unless otherwise specified.
Dataset-specific details, including the dissimilarity function $s(\cdot)$, inclusion of Dice loss, and other training parameters, are provided in the appendix.
For a fair comparison, all models were trained under identical conditions or, where necessary, using the optimal settings outlined in their original publications.

\subsubsection{Baseline Methods}
We compare our EOIR framework with several state-of-the-art non-iterative, learning-based baseline models, including VoxelMorph \cite{balakrishnan2019voxelmorph}, TransMorph \cite{chen2022transmorph}, LKU-Net \cite{jia2022u}, Fourier-Net \cite{jia2023fourier}, RDP~\cite{wang_RDP}, LapIRN \cite{mok2021large}, MemWarp \cite{zhang2024memwarp} and CorrMLP \cite{meng2024correlation}. 
For the Abdomen CT dataset, we additionally include discrete optimization-based methods ConvexAdam \cite{siebert2021fast} and SAMConvex \cite{li2023samconvex} in the comparison, as these are highly effective for handling large deformations. 
For multi-modality 2D image registration, SOTA 2D multi-modality image registration approaches like PGMR \cite{zheng2025plug} and IMF\cite{wang2024improving} are also compared.
For results on the OASIS and LUMIR datasets, we obtained evaluation scores from the public leaderboard or respective publications. 
In the case of the ACDC, Abdomen CT, HippocampusMR, and RGB-IR datasets, we used publicly available code for each model and fine-tuned them to achieve optimal performance.

\subsubsection{Evaluation Metrics}
Following established methodologies \cite{dalca2018unsupervised,balakrishnan2019voxelmorph,mok2021large,chen2022transmorph} and challenge protocols \cite{hering2022learn2reg}, we evaluate anatomical alignment using the Dice Similarity Coefficient (Dice) and the 95\% Hausdorff Distance (HD95). 
Target registration error (TRE) is also computed if the dataset provides ground-truth landmark points.
For 2D multi-modality image registration, normalized mutual information (nMI), normalized cross-correlation (NCC),
and peak signal-to-Noise ratio (PSNR) are utilized, following \cite{zheng2025plug}.
To assess field smoothness, we measure the standard deviation of the logarithm of the Jacobian determinant (SDlogJ), and non-diffeomorphic volumes (NDV) \cite{liu2024finite}. 
Additionally, computational complexity is assessed using the multiply-add operations (Multi-Adds, G) and total parameter size (Params, MB). The inference time on cardiac image registration and hippocampus image registration are also presented to demonstrate the efficiency of EOIR.

\subsection{Results \& Analysis on Registration Accuracy}

\subsubsection{Handling Large Deformations}
\begin{figure*}[h]
    \centering
    \includegraphics[width=1.0\linewidth]{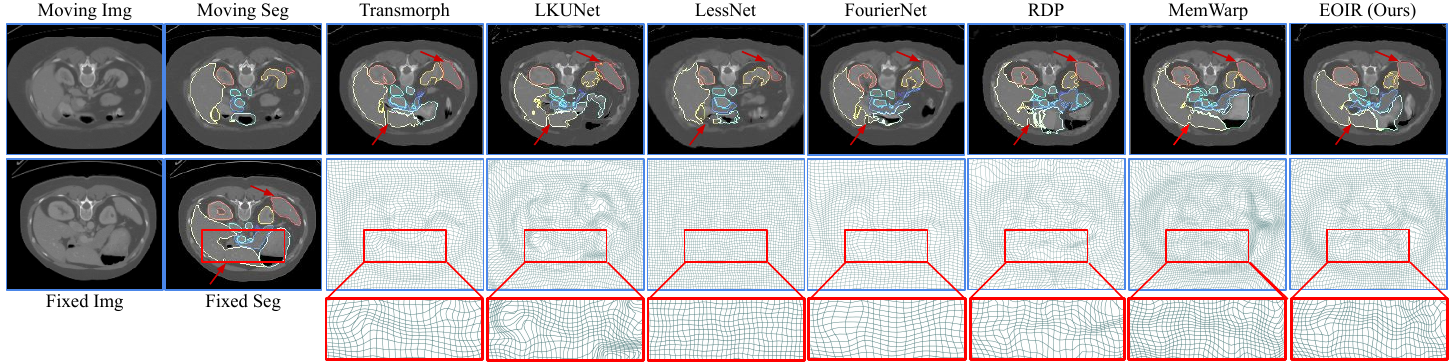} 
    \caption{
    \textcolor{resp}{Qualitative comparison on the abdomen CT dataset. 
    TransMorph, LessNet, and FourierNet exhibit smooth deformation fields but fall short in anatomical alignment. 
    MemWarp and LKUNet improve alignment but introduce more implausible voxels, while EOIR shows better accuracy-smoothness balance.}}
    
    \label{fig:qualitative_abdomen}
    \vspace{-2ex}
\end{figure*}

We demonstrate the capability of handling large deformations using inter-subject abdominal CT registration. 
As shown in Table \ref{tab:abdomen} and Fig. \ref{fig:qualitative_abdomen}, EOIR outperforms all other methods in registration accuracy without compromising smoothness.
Specifically, EOIR surpasses the best conventional learning-based method, LKU-Net, by 14.87\%, the best image pyramid-based method, MemWarp, by 0.65\%, the best efficiency-driven method, FourierNet, by 41.66\%, and the best discrete optimization-based method, SAMConvex, by 13.01\%.
It's worth noting that while MemWarp matches EOIR in registration accuracy, it uses additive deformation composition, resulting in a less smooth field. 
Although FourierNet achieves a smoother field, it falls behind in registration accuracy. 
Both discrete optimization-based methods generate smooth deformation fields and improve upon efficiency-driven methods, but their registration accuracy lags behind all image pyramid-based methods and EOIR.

\begin{table}[!h]
\begin{center}
\caption{
Quantitative comparison on the abdomen CT dataset. 
Best-performing metrics are in bold. 
Symbols indicate direction: $\uparrow$ for higher is better, $\downarrow$ for lower is better. 
``Initial" refers to baseline results before registration. 
}
\label{tab:abdomen}
\resizebox{0.98\columnwidth}{!}{
\begin{tabular}{ lccc }
\hline
\hline
\rowcolor{lightgray}
Model &  Dice (\%) $\uparrow$ & HD95 (mm) $\downarrow$ & SDlogJ $\downarrow$ \\ 
\hline
Initial & 30.86 & 29.77 & - \\
\hline
VoxelMorph \cite{balakrishnan2018unsupervised} & 47.05 & 23.08 & 0.13 \\
TransMorph \cite{chen2022transmorph} & 47.94 & 21.53 & 0.13 \\
LKUNet \cite{jia2022u} & 52.78 & 20.56 & 0.98 \\
\hline
LapIRN \cite{mok2020large} & 54.55 & 20.52 & 1.73 \\
CorrMLP \cite{meng2024correlation} & 56.11 & 19.52 & 0.16 \\
RDP \cite{wang_RDP} & 58.77 & 20.07 & 0.22 \\
MemWarp \cite{zhang2024memwarp} & 60.24 & 19.84 & 0.53 \\
\hline
LessNet \cite{jia2024decoder} & 42.03 & 27.03 & \bf{0.07} \\
FourierNet \cite{jia2023fourier} & 42.80 & 22.95 & 0.13 \\
\hline
ConvexAdam \cite{siebert2021fast} & 51.10 & 23.14 & 0.11\\ 
SAMConvex \cite{li2023samconvex} & 53.65 & 18.66 & 0.12 \\
VoxelOpt \cite{zhang2025voxelopt} & 58.51 & 18.54 & 0.21 \\
\hline
uniGradICON \cite{uniGradICON2024MICCAI} & 53.33  & 20.20 & 0.13 \\
\hline
\textbf{EOIR (-MS loss)} & 59.57 & 18.98 & 0.29 \\
\textbf{EOIR (-M3)} & 60.12 & 18.46 & 0.17 \\
\textbf{EOIR (-M4)} & 59.60 & 19.04 & 0.17 \\
\textbf{EOIR (Ours)} & \bf{60.63} & \bf{17.61} & 0.17 \\
\hline
\hline
\end{tabular}
}
\end{center}
\vspace{-2ex}
\end{table}

\subsubsection{Handling Local Intra-subject Motions}
Unlike inter-subject abdominal CT registration, intra-subject cardiac registration focuses on tracking local cardiac motion, such as the movement of the left and right ventricles, during the complete phases of the cardiac cycle.
As shown in Table \ref{tab:ACDC} and Fig. \ref{fig:3data}, EOIR outperforms all other methods in registration accuracy, measured by Dice score. 
Specifically, EOIR surpasses the best conventional learning-based method, LKU-Net, by 3.1\%, the best image pyramid-based method, RDP, by 1.1\%, and the best efficiency-driven method, FourierNet, by 3.0\%. 
It is worth noting that LessNet and LapIRN are excluded from Table \ref{tab:ACDC}, as they cannot handle short-axis data, and no straightforward approach was found to enable them to do so. With only 7.2\% MAs and 1.2\% parameters, EOIR($Ns=8$) matches RDP's performance. This efficiency is crucial for deployment on resource-limited hardware.

\begin{table}[!h]
\begin{center}
\caption{
Quantitative comparison on the cardiac ACDC dataset. 
Best-performing metrics are highlighted in bold. 
``MAs (G)" stands for multi-adds (G), and ``PS (MB)" is parameter size (MB).
``Time'' is the average inference time for 1 registration pair. GM(MB) is the occupied GPU memory in inference.
}\label{tab:ACDC}
\resizebox{1.\linewidth}{!}{
\begin{tabular}{ lcccrrcc }
\hline
\hline
\rowcolor{lightgray}
Model & Dice (\%) $\uparrow$ & HD95 (mm) $\downarrow$ & SDlogJ $\downarrow$ & MAs (G) $\downarrow$ & PS (MB) $\downarrow$ & Time $\downarrow$ & GM(GB) $\downarrow$\\ 
\hline
Initial & 58.14 & 11.95 & - & -~~~~~~ & -~~~~~~ & - & -\\
\hline
TransMorph \cite{chen2022transmorph} & 74.97 & 9.44 & 0.045 & 50.20~~~ & 46.69~~~ & 0.26 & 18.3\\
VoxelMorph \cite{balakrishnan2018unsupervised} & 75.26 & 9.33 & \textbf{0.044} & 19.55~~~ & 0.32~~~ & \textbf{0.18} & \textbf{2.7}\\
LKU-Net \cite{jia2022u} & 76.53 & 9.13 & 0.049 & 160.50~~~ & 33.35~~~ & 0.22 & 3.9\\
\hline
Fourier-Net \cite{jia2023fourier} & 76.61 & 9.25 & 0.047 & 86.07~~~ & 17.43~~~ & 0.27 & 3.1\\
\hline
CorrMLP \cite{meng2024correlation} & 77.31 & \textbf{9.00} & 0.056 & 47.59~~~ & 4.19~~~ & 0.28 & 3.3\\
MemWarp \cite{zhang2024memwarp} & 76.74 & 9.67 & 0.108 & 1270.00~~~ & 47.78~~~ & 0.58 & 12.7\\
RDP \cite{wang_RDP} & 78.06 & 9.02 & 0.076 & 154.00~~~ & 8.92~~~ & 0.36 & 4.1\\
\hline
EOIR ($N_s=8$) & 78.28 & 9.14 & 0.071 & 11.02~~~ & 0.11~~~ & 0.25 & 2.9\\
\textbf{EOIR (Ours)} & \textbf{78.91} & 9.07 & 0.084 & 114.21~~~ & 0.91~~~ & 0.26 & 4.5\\
\hline
\hline
\end{tabular}
}
\end{center}
\vspace{-2ex}
\end{table}

\begin{table}[!h]
\begin{center}
\caption{
Quantitative comparison on the OASIS dataset. 
``TransMorph-1" and ``TransMorph-2" denote versions with different smoothness regularization.  
}
\label{tab:OASIS}
\resizebox{0.98\columnwidth}{!}{
\begin{tabular}{ lccc }
\hline
\hline
\rowcolor{lightgray}
Model &  Dice (\%) $\uparrow$ & HD95 (mm) $\downarrow$ & SDlogJ $\downarrow$ \\ 
\hline
Initial & 57.18 & 3.83 & - \\
\hline
VoxelMorph \cite{balakrishnan2018unsupervised} & 84.70 & 1.55 & 0.13 \\
TransMorph-1 \cite{chen2022transmorph} & 86.20 & 1.43 & 0.13 \\
TransMorph-2 \cite{chen2022transmorph} & 88.54 & 1.27 & 0.50 \\
LKUNet \cite{jia2022u} & 88.61 & \textbf{1.26} & 0.52 \\
\hline
LessNet \cite{jia2024decoder} & 78.80 & 2.15 & 0.10 \\
FourierNet \cite{jia2023fourier} & 86.04 & 1.37 & 0.48 \\
\hline
LapIRN \cite{mok2020large} & 86.10 & 1.51 & \textbf{0.07} \\
\hline
ConvexAdam~\cite{siebert2021fast} & 84.64 & 1.50 & \textbf{0.07} \\
\hline
EOIR ($N_s$=16) & 86.96 & 1.38 & 0.28\\
\textbf{EOIR (Ours)} & \textbf{88.83} & 1.28 & 0.52 \\
\hline
\hline
\end{tabular}
}
\end{center}
\vspace{-2ex}
\end{table}

\begin{table}[!h]
\begin{center}
\caption{
Quantitative comparison on the LUMIR dataset. 
``EOIR (addition)" refers to using addition to compose fields.
}
\label{tab:LUMIR}
\resizebox{0.98\columnwidth}{!}{
\begin{tabular}{ lcccc }
\hline
\hline
\rowcolor{lightgray}
Model &  Dice (\%) $\uparrow$ & HD95 (mm) $\downarrow$ & NDV (\%) $\downarrow$ & TRE (mm) $\downarrow$\\ 
\hline
Initial & 56.57 & 4.79 & - & -\\
\hline
DeedsBCV \cite{heinrich2013mrf} & 69.77 & 3.95 & 0.000 & \textbf{2.22}\\
SynthMorph \cite{hoffmann2021synthmorph} & 72.43 & 3.57 & 0.000 & 2.61\\
VoxelMorph \cite{balakrishnan2018unsupervised} & 73.25 & 3.76 & 0.397 & 2.69\\
TransMorph \cite{chen2022transmorph} & 75.94 & 3.51 & 0.351 & 2.42\\
\hline
EOIR (addition) & 77.34 & 3.34 & 0.186 & 2.37\\
\textbf{EOIR (Ours)} & \textbf{77.37} & \textbf{3.33} & \textbf{0.000} & 2.35\\
\hline
\hline
\end{tabular}
}
\end{center}
\vspace{-2ex}
\end{table}

\subsubsection{Handling Brain MR Image Registration}
Unlike cardiac and abdomen datasets with different organ motions, inter-subject brain MR image registration requires fine-grained alignments of multiple variably shaped and sized brain structures. 
Table \ref{tab:OASIS} presents the results of semi-supervised learning on the OASIS dataset. 
With similar field smoothness, EOIR outperforms both LKUNet and TransMorph-2 in Dice score. 
Note that these models were trained semi-supervisedly, and the results are clearly influenced by the provided segmentation masks during training, as improvements in Dice score often correlate with less feasible deformation fields.
The advantage of EOIR in this task primarily lies in the reduction of parameter size. 
With similar smoothness and a slightly higher Dice score, EOIR reduces network parameter size by 98.1\% compared to TransMorph (46.69 MB) and by 97.3\% compared to LKUNet (33.35 MB) (the network parameter sizes are the same as those in Table \ref{tab:ACDC}).
Therefore, we further evaluate EOIR on the LUMIR dataset, which is large-scale and trained in an unsupervised manner.
As shown in Table \ref{tab:LUMIR}, EOIR outperforms TransMorph by 1.9\% with the NDV close to zero.
Additionally, while SynthMorph \cite{hoffmann2021synthmorph} and DeedsBCV \cite{heinrich2013mrf} provide diffeomorphic deformation fields, their anatomical alignment falls short of EOIR. 
EOIR improves Dice by 6.8\% and 10.9\%, respectively, compared to SynthMorph and DeedsBCV. Compared to direct addition (EOIR(addition)), EOIR's composition approach delivers folding-free deformation fields (0\% NDV) and superior registration performance with only marginal computational overhead, underscoring the strategy of deformation composition for high-quality image registration.

\subsubsection{Handling Structures with Ambiguous Boundaries}  
\begin{figure*}[htbp]
    \centering
    \includegraphics[width=1.0\linewidth]{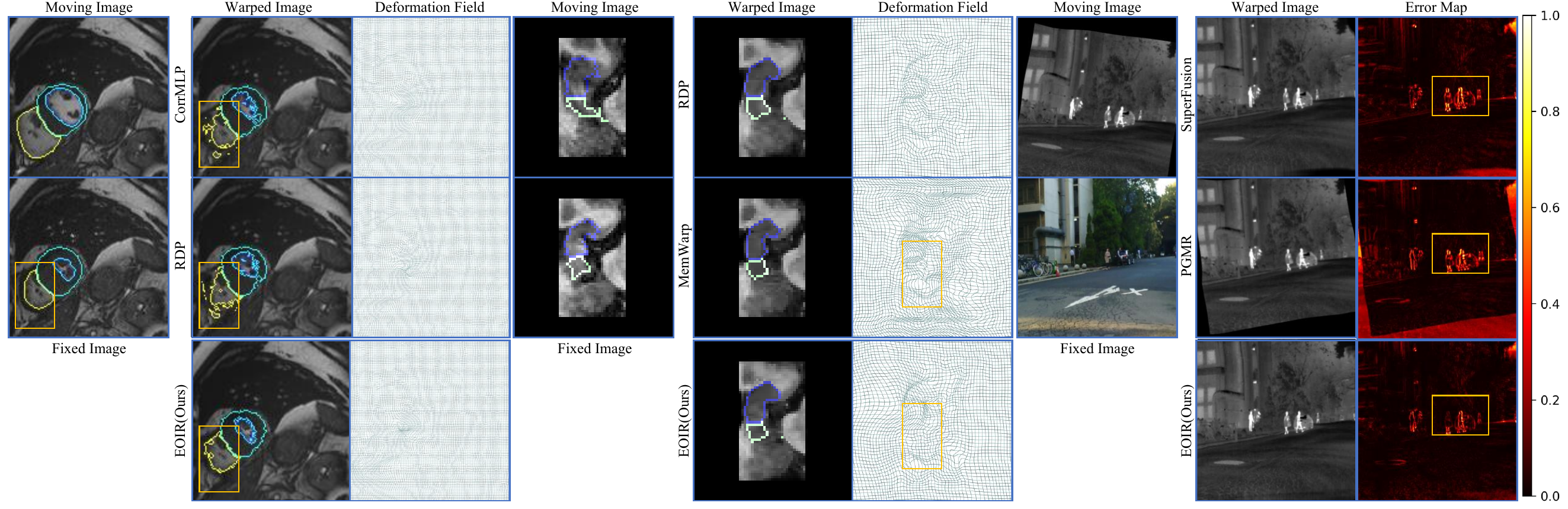} 
    \caption{
    \textcolor{resp}{Qualitative comparison on ACDC, HippocampusMR, and RGB-IR datasets (from left to right, respectively). For the results on each dataset, we compare EOIR with two sub-optimal approaches. The Error map is presented for RGB-IR dataset (color bar on the right).}
    }
    \label{fig:3data}
    \vspace{-2ex}
\end{figure*}

The HippocampusMR dataset involves aligning two neighboring structures (the hippocampus head and body), with less defined boundaries compared to other anatomical regions.  
Results on this dataset are summarized in Table~\ref{tab:HippocampusMR} and Fig. \ref{fig:3data}.  
Our EOIR framework consistently outperforms other approaches, except MemWarp \cite{zhang2024memwarp}.  
MemWarp, designed for discontinuity-preserving registration, jointly predicts segmentation masks and deformation fields, enforcing stronger regularization on structural boundaries.  
While MemWarp achieves marginally higher Dice scores, its deformation fields are significantly less smooth than those produced by EOIR.  
This demonstrates EOIR’s superior balance between registration accuracy and deformation smoothness.

\begin{table}[!h]
\begin{center}
\caption{
Quantitative comparison on the HippocampusMR dataset. 
}
\label{tab:HippocampusMR}
\resizebox{0.98\columnwidth}{!}{
\begin{tabular}{ lcccc }
\hline
\hline
\rowcolor{lightgray}
Model &  Dice (\%) $\uparrow$ & HD95 (mm) $\downarrow$ & SDlogJ $\downarrow$ & Time $\downarrow$ \\ 
\hline
Initial & 62.46 & 12.39 & - \\
\hline
VoxelMorph \cite{balakrishnan2018unsupervised} & 80.97 & 7.90 & 0.06 & \textbf{0.19}\\
TransMorph \cite{chen2022transmorph} & 84.90 & 6.19 & 0.07 & 0.49\\
RDP \cite{wang_RDP} & 86.13 & 6.02 & 0.07 & 0.46\\
CorrMLP \cite{meng2024correlation} & 84.86 & 6.55 & 0.07 & 0.29\\
LessNet \cite{jia2024decoder} & 75.59 & 9.28 & 0.05 & 0.30\\
FourierNet \cite{jia2023fourier} & 80.10 & 7.59 & 0.06 & 0.33\\
LKUNet \cite{jia2022u} & 75.09 & 9.29 & \textbf{0.04} & 0.39\\
MemWarp \cite{zhang2024memwarp} & \textbf{86.50} & \textbf{5.71} & 0.24 & 0.96\\
\hline
\textbf{EOIR (Ours)} & 86.44 & 6.20 & 0.10 & 0.35\\
\hline
\hline
\end{tabular}
}
\end{center}
\vspace{-2ex}
\end{table}

\subsubsection{Handling 2D Multi-modality Natural Images}

In addition to medical images, we also demonstrated our EOIR on 2D natural images, using images from the RGB-IR dataset \cite{tang2022piafusion}. As shown in Table \ref{tab:RGB-IR} and Fig. \ref{fig:3data}, despite the modality difference between the moving and fixed images, EOIR can still obtain comparable registration performance to the state-of-the-art approaches in natural image registration (PGMR and Superfusion), highlighting the robustness of our EOIR in handling natural images. Notably, it accomplished this with substantially fewer parameters (reduced 78\% of parameters compared with Superfusion), demonstrating both the robustness and efficiency of our framework. We posit that EOIR's effectiveness stems from the high texture and structural similarity between RGB and IR images, even though they differ in modality. For more complex multimodal scenarios, incorporating a more powerful encoder may be necessary to further enhance performance.

\begin{table}[!h]
\begin{center}
\caption{
Quantitative comparison on the RGB-IR dataset.
}
\label{tab:RGB-IR}
\begin{tabular}{ lcccc }
\hline
\hline
\rowcolor{lightgray}
Model &  nMI (\%) $\uparrow$ & NCC (\%) $\uparrow$ & PSNR $\uparrow$ & PS (MB) $\downarrow$\\ 
\hline
Initial & -& 47.52 & 22.96& -\\
\hline
SIFT \cite{lowe2004distinctive} & 36.97 & 42.63 & 23.10 & - \\
ReCoNet~\cite{huang2022reconet} & 93.40 & 63.71 & 25.01 & 3.09 \\
Superfusion~\cite{tang2022superfusion} & 95.29 & 87.70 & 29.98 & 6.96 \\
IMF~\cite{wang2024improving} & 93.95 & 74.77 & 26.42 & 27.13 \\
PGMR~\cite{zheng2025plug} & \textbf{95.65} & 87.61 & 29.55 & 1670.16\\
\hline
\textbf{EOIR (Ours)} & 95.17 & \textbf{88.17} & \textbf{30.23} & \textbf{1.50}\\
\hline
\hline
\end{tabular}
\end{center}
\vspace{-2ex}
\end{table}

\subsection{Results \& Analysis on Computational Complexity}
In this section, we provide results and analysis of EOIR regarding computational complexity. 
Two key parameters influence EOIR's complexity: the number of convolutional layers \(n_c\) in the encoder, and the start channel \(N_s\), as depicted in Fig. \ref{fig:encoder_flow_estimator}.
To provide a more striking comparison, we conduct complexity analysis using the most challenging abdomen dataset, alongside the ACDC dataset, which features much smaller image sizes compared to the other three datasets.
\subsubsection{Effects of $n_c$}
As shown in Fig. \ref{fig:nconvs_dsc}, increasing $n_c$ generally improves registration accuracy. 
However, starting from $n_c=3$, the marginal benefits diminish rapidly with each additional layer. 
Interestingly, when replacing the encoder with a U-Net, the accuracy decreases despite the increased parameter size. 
Sub-optimal methods like RDP, utilizing a similar U-Net architecture for pyramid registration, lag behind EOIR in all metrics. 
This suggests that mono-modal registration benefits more from local intensity linearization than from global bias harmonization, unlike affine registration \cite{wang2023robust}.

\begin{figure}[ht]
    \centering
    \includegraphics[width=0.9\linewidth]{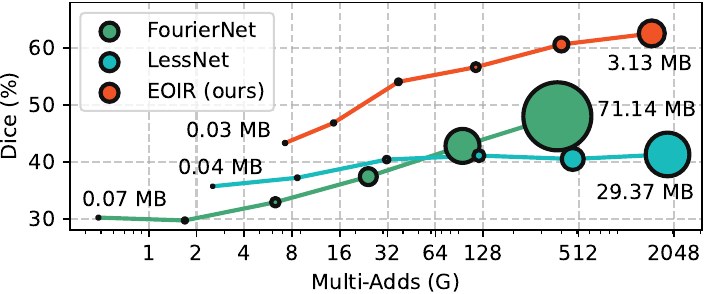}
    \caption{Visual comparison of the trade-off between Dice and computational complexity for varying start channels ($N_s$ from 2 to 64, doubling each step), compared to top-performing efficiency-driven methods on the Abdomen CT dataset. 
    Circle size and label indicate network parameter size.}
    \label{fig:barchat}
    \vspace{-2ex}
\end{figure}

\subsubsection{Effects of $N_s$}
Keeping three conv layers intact, we further study the impact of start channels \(N_s\) on the outcome.
We compare EOIR with efficiency-driven methods, FourierNet and LessNet.
As shown in Fig.\ref{fig:barchat}, starting from \(N_s=2\) for all methods, EOIR shows a much slower increase in parameter size compared to the other two.
While both FourierNet and EOIR exhibit a log-linear increase in accuracy relative to multi-adds and parameter size, EOIR achieves a much better accuracy-efficiency balance than FourierNet.
When assessed on the ACDC dataset, as shown in Table \ref{tab:ACDC}, EOIR's improvement is more pronounced.
EOIR ($N_s=8$) surpasses VoxelMorph, the runner-up in terms of complexity, with a significant increase in Dice score while maintaining lower multi-adds and parameter size.
Also, EOIR ($N_s=8$) reduces the parameter size by 98.8\% and multi-adds by 92.9\% compared to RDP, the runner-up in accuracy, while achieving a slightly higher Dice score.

\begin{figure}[ht]
    \centering
    \includegraphics[width=1.0\linewidth]{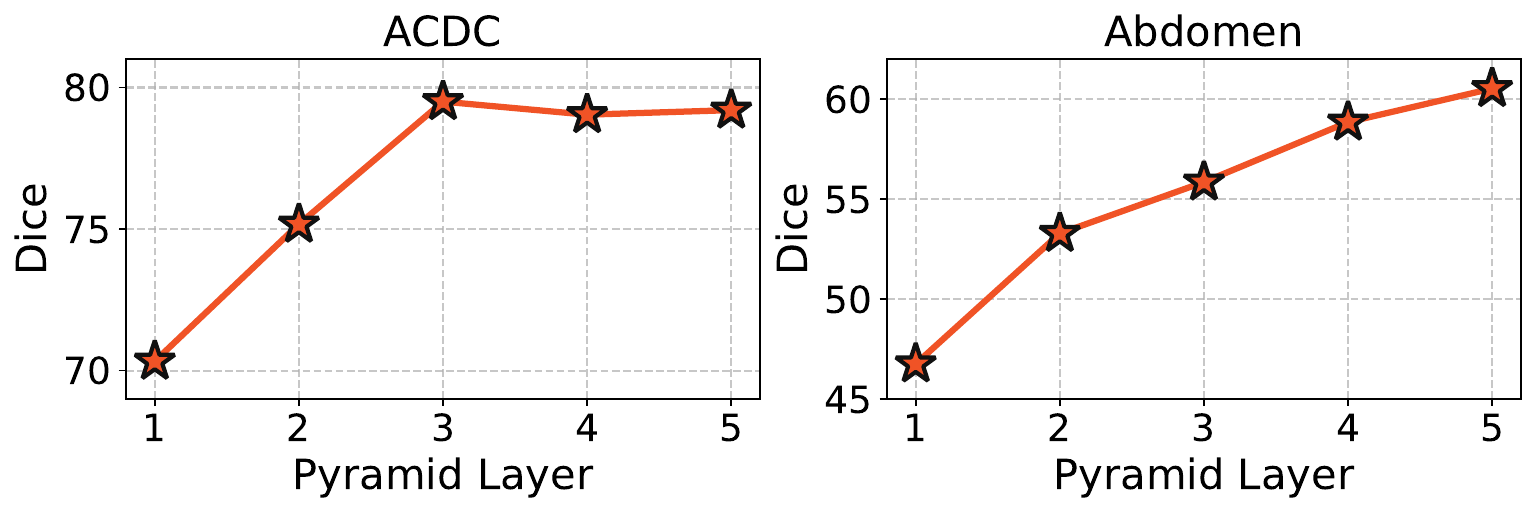}
    \caption{
    Dice score with the increase of pyramid layers in ACDC and Abdomen images.}
    \label{fig:dice_pyramid}
    \vspace{-2ex}
\end{figure}

\subsubsection{Effects of Pyramid Layers $n$}
\label{sec:results_effects_pyramid_n}
The number of pyramid layers $n$ determines the ability of EOIR to capture large deformation. 
We plot the curve of the registration Dice with the increasing pyramid layers, as shown in Fig.~\ref{fig:dice_pyramid}. 
For intra-patient registration on ACDC, three pyramid layers are sufficient to capture the ED-ES or ES-ED deformation. For inter-patient registration on abdomen images, the registration Dice keeps increasing from 1-5 pyramid layers. Therefore, to get the optimal number of pyramid layers, priors in the registration task should be considered, while a five-layer pyramid can work for most registration tasks. Note that, the multi-scale supervision strategy contributes significantly to accuracy and smoothness: ablating it in favor of a single-scale loss on the top layer alone causes a 1\% Dice drop (with SDlogJ from 0.17 to 0.29) in abdominal image registration (Table \ref{tab:abdomen}, EOIR(-MS loss)). Furthermore, to proactively improve the model's inherent capability to handle large deformations, the training process itself could be fortified by using a pyramid with an extra level, ensuring that EOIR learns a more robust feature representation across an even wider range of motion.


\subsection{Results \& Analysis on Smoothness \& Diffeomorphism}
In this section, we present the results and analysis of EOIR on deformation field smoothness.
For the unsupervised setting, we evaluate the ACDC and LUMIR datasets, while for the semi-supervised setting, we focus on the Abdomen CT and OASIS datasets.
We emphasize the use of Eq.~\eqref{eq:xia_ji_ba_xie} for composing deformation fields across pyramid levels and time intervals.
With $n \times m$ sufficiently small time intervals, each deformation $\phi^{ij}$ can be considered diffeomorphic.
The composition is performed by resampling one deformation field by another, ensuring that the resulting deformation remains diffeomorphic, as described in Eq.~\eqref{eq:xia_ji_ba_xie}.

\subsubsection{Unsupervised}
\begin{figure}[ht]
    \centering
    \includegraphics[width=1.0\linewidth]{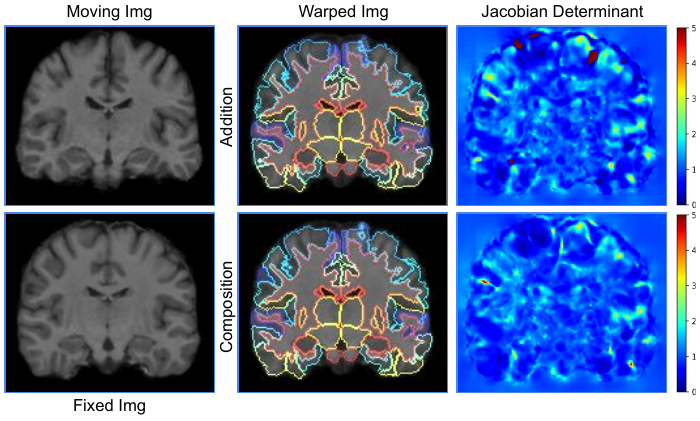}
    \caption{
        Qualitative comparison between addition-based and composition-based registration in LUMIR. For the Jacobian determinants of deformation fields, Jacobian determinants $<0$ are highlighted in dark red. 
    }
    \label{fig:qualitative_lumir}
    \vspace{-2ex}
\end{figure}

As shown in Table \ref{tab:ACDC}, all methods, except MemWarp (SDlogJ: 0.11), produced smooth outputs with SDlogJ values below 0.10. 
Notably, EOIR ($N_s=8$) achieves a lower SDlogJ while delivering a slightly better Dice score than the top-performing pyramid-based method, RDP, demonstrating EOIR's superior handling of local motions during the cardiac cycle.
In brain MR registration, the effect of using the proposed composition method in Eq.~\eqref{eq:xia_ji_ba_xie} becomes more evident. 
As seen in Table \ref{tab:LUMIR}, additive composition results in significantly more non-diffeomorphic voxels for EOIR, whereas applying Eq.~\eqref{eq:xia_ji_ba_xie} reduces NDV (\%) to nearly zero without sacrificing accuracy (see Fig. \ref{fig:qualitative_lumir} for a visual example). 
For other baseline methods, improvements in anatomical alignment are often accompanied by an increase in NDV (\%).

\subsubsection{Semi-supervised}
\begin{figure}[ht]
    \centering
    \includegraphics[width=0.9\linewidth]{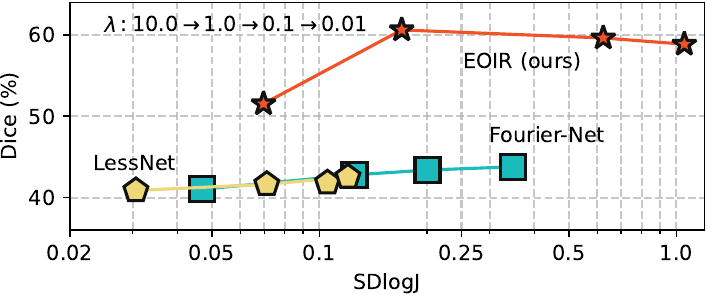}
    \caption{Visual comparison of the trade-off between Dice and smoothness (SDlogJ) for varying $\lambda$ in Eq.~\eqref{eq:loss} (0.1 to 10.0, increasing $10\times$ per step), compared to efficiency-driven methods on the Abdomen CT dataset. 
    } 
    \label{fig:dice_smooth}
    \vspace{-2ex}
\end{figure}

Semi-supervised learning often incorporates Dice loss to promote anatomical alignment, with the resulting deformation field typically being influenced by the segmentation masks.
This can sometimes lead to implausible deformation fields, as the optimization is driven more by the mask alignment than the underlying image data.
As shown in Table \ref{tab:OASIS}, methods including EOIR demonstrate that better anatomical alignment, indicated by higher Dice and lower HD95 (mm), often results in a larger SDlogJ, signaling a less smooth deformation field.
For TransMorph, decreasing smoothness regularization leads to increasing Dice and decreasing SDlogJ.
Similarly, for EOIR, by simply reducing the start channel to $N_s=16$, a similar accuracy-smoothness trade-off is observed.
Abodmen CT dataset is more challenging as it has large deformation and limited training samples, and the accuracy-smoothness variations are larger than the OASIS.
Therefore, we further study how smoothness regularization strength affect EOIR and other efficiency-driven methods.
As shown in Fig. \ref{fig:dice_smooth}, for LessNet and FourierNet, decreasing $\lambda$ reduces field smoothness while increasing Dice. 
In contrast, EOIR exhibits a distinct behavior due to its use of Eq.~\eqref{eq:xia_ji_ba_xie} for large-deformation diffeomorphic transformation, which imposes stricter requirements on field plausibility at each pyramid level. For EOIR, we found that decreasing $\lambda$ initially increases the Dice score, which peaks at $\lambda=1.0$ before declining. We therefore empirically set $\lambda=1.0$ as it represents the optimal trade-off.
Additionally, varying $n_c$ and $N_s$ changes Dice, but SDlogJ remains around $0.17$, highlighting the effectiveness of Eq.~\eqref{eq:xia_ji_ba_xie} in handling large deformations (see Fig. \ref{fig:qualitative_abdomen}).


\subsection{Ablation Study on Modifications}

We performed a systematic ablation study to evaluate key components of EOIR using abdominal CT data (Table \ref{tab:abdomen}). Our separable feature design (\textbf{M1}) proved fundamental compared to combined feature methods, while pyramid analysis (\textbf{M2}, Fig. \ref{fig:dice_pyramid}) confirmed the value of multi-scale processing. Removing the Hadamard transform (\textbf{M3}) impaired registration accuracy, and omitting Gaussian smoothing in the multi-scale loss (\textbf{M4}) substantially degraded performance by destabilizing gradient propagation. Optimal efficiency-accuracy balance was achieved with a three-layer encoder (\textbf{M5}, Fig. \ref{fig:nconvs_dsc}), validating our architectural configuration.

\subsection{Discussion}

\subsubsection{Summary} 
Overall, EOIR achieves a superior balance between accuracy-efficiency and accuracy-smoothness in deformable image registration. 
Across five datasets with varying modalities and anatomies, EOIR reduces computational complexity without compromising accuracy. 
Moreover, the novel deformation field composition method in Eq.~\eqref{eq:xia_ji_ba_xie} enables EOIR to maintain smoother deformation fields while preserving accuracy.
The performance gains of EOIR stem from its integration of the Horn–Schunck (H-S) assumption and the Linearization-Harmonization (L-H) assumption into the network design. 
We propose that a few convolutional layers are sufficient for feature extraction using a Laplacian feature pyramid, effectively linearizing local intensities and harmonizing mono-modal images at each pyramid level. 
The number of voxels satisfying the H-S assumption in both moving and fixed feature maps increases at each level, promoting displacement propagation within the smoothness regularization.

\subsubsection{Advantages}
Two major advantages emerge with the design of EOIR. 
\textbf{Expansibility:} EOIR's simple architecture, consisting of just a few convolutional layers, makes it highly adaptable to incorporate advanced network modules such as large-kernel convolutions, transformer blocks, co-attention, and other novel structures. 
For instance, replacing the 3-layer ConvNet encoder with a full encoder-decoder U-net can enhance feature extraction capabilities.  
Additionally, incorporating larger-kernel convolutions or self-attention modules after the 3-layer ConvNet flow estimator can expand the network's effective receptive field.  
\textbf{Efficiency}: EOIR's efficient design significantly reduces computational complexity without sacrificing accuracy, making it well-suited for large volumetric images and deployment in resource-constrained settings.
Specifically, Figure~\ref{fig:barchat} illustrates the trade-off between Dice and computational complexity, measured by Multi-Adds (G) and network parameter size.
Runtime comparisons among methods are shown in Table~\ref{tab:ACDC} and Table~\ref{tab:HippocampusMR}, demonstrating that all these learning-based methods can perform fast forward inference for volumetric registration within one second.

\subsubsection{Zero-shot Inference Analysis}
Beyond the results presented in this manuscript, the zero-shot inference capability of our EOIR framework has been further substantiated across multiple additional image registration tasks, as reported in \cite{chen2025lumir} and the Appendix (under the team name `next-gen-nn'). Specifically, EOIR consistently ranks among the top three methods in registration accuracy for inter-subject registration using both the ADNI-1.5T and ADNI-3T datasets. It demonstrates particular strength in subject-to-atlas registration, achieving first place on the NIMH-T1w dataset \cite{nugent2022nimh} and second place on both the ADHD \cite{lytle2020neuroimaging,lytle2021neuroimaging} and UltraCortex-9.4T \cite{mahler2024ultracortex} datasets. Across all these evaluations, EOIR achieves the lowest NDV among top-ranked approaches, underscoring its capacity to maintain diffeomorphic properties without compromising registration accuracy. Notably, EOIR is the only top-tier method that does not employ a progressive registration strategy, as highlighted in \cite{chen2025lumir}. We posit that this advantage stems from the incorporation of the H–S and L–H assumptions into the network architecture, key elements overlooked by other methods. Their omission results in a less favorable balance between accuracy and computational efficiency, reinforcing EOIR's value as an elegant and highly effective registration framework.

\subsubsection{Broader Impact}
Beyond benchmarking accuracy and efficiency, EOIR has broader implications for critical neuroimaging applications that demand precise and stable voxel-to-voxel alignment.
\textbf{Longitudinal image alignment:} Accurate deformable registration is fundamental for tracking intra-subject structural changes over time, such as lesion progression in multiple sclerosis or tumor evolution in oncology. 
By accurately aligning longitudinal images with smooth and efficient deformation fields that account for subject-specific brain atrophy, EOIR improves sensitivity to subtle lesion growth, shrinkage, or transformation, enabling reliable lesion- or tumor-level quantification across time points.
This is exemplified in our recent work on longitudinal unique lesion tracking (AULTRA)~\cite{zhang2025aultra}, which also leverages our advances in MS lesion segmentation~\cite{zhang2025uniself} and filling~\cite{zhang2025bi} as preprocessing for EOIR-based registration, followed by unique lesion identification~\cite{rivas2025unique} across time points.
\textbf{Atlas construction:} Many population-level analyses, including those in quantitative susceptibility mapping (QSM)~\cite{wang2015quantitative}, rely on accurate voxel-wise correspondences across individuals. 
A future direction is to leverage EOIR for constructing susceptibility and multimodal brain atlases with accurate voxel-wise alignment to study iron deposition, myelin integrity, and other quantitative biomarkers across the cohort.

\subsubsection{Limitations}  
EOIR has two main limitations.  
First, while EOIR achieves high accuracy and efficiency in mono-modal registration, its performance on multi-modal data is less effective.  
This occurs because the three-layer convolutional encoder, designed to linearize intensities and harmonize local contrast, lacks the capacity to learn cross-modal feature invariance \cite{wang2023robust}, which requires deeper networks to capture larger contextual regions.  
For example, when applied to the ThoraxCBCT dataset \cite{hugo2017longitudinal} (registering pre-therapeutic FBCT to low-dose CBCT), EOIR achieves a Dice score of 45\%. Replacing the encoder with a full U-Net (while retaining the EOIR framework) increases Dice to 56\%, validating the framework’s generalizability but highlighting the three-layer encoder’s trade-off between harmonization capacity and efficiency (details can be found in Table XII in the Appendix). For complex multi-modality image registration, EOIR can also incorporate a more powerful encoder (e.g. encoder from foundation models) to tackle this challenge.

Second, EOIR shows diminished effectiveness when processing images that have fine-grained features, for example, lung nodules or retinal vessels.
The reason lies in the feature pyramid design, which incorporates spatial downsampling. 
During this downsampling procedure, small structures like lung nodules or retinal vessels can be gradually reduced in prominence or even lost.
Therefore, at certain pyramid level $i$, the feature pyramid scheme is likely to miss objects with a largest diagonal dimension smaller than \(2^{(i - 1)}\) voxels.
Even if the displacement of these small objects may exceed their own size, this results in relatively poorer performance compared to cascaded methods.

\section{Conclusion}
In this paper, we introduced EOIR, a simple yet efficient image registration network that departs from conventional learning-based approaches by eliminating the decoder and relying solely on an encoder for feature extraction. 
This streamlined design substantially reduces the number of parameters compared to traditional methods. 
Extensive experiments across 10 datasets (6 main datasets+validation on 4 datasets) demonstrate that EOIR not only achieves notable improvements in registration accuracy but also reduces network complexity, compared to state-of-the-art methods. 
EOIR effectively handles both small and large deformations through a novel deformation composition scheme, striking a balance between accuracy, efficiency, and smoothness. 
With lightweight design and strong performance, EOIR serves as a robust backbone for future developments in more complex registration architectures, offering a solid foundation for scaling in resource-constrained or large volumetric settings.



\section*{Acknowledgments}
The authors would like to express their sincere gratitude to Aaron Carass for his invaluable insights and constructive comments, which significantly improved the manuscript's quality. 

\bibliographystyle{ieeetr}
\bibliography{eoir}

@inproceedings{zhang2023spatially,
  title={Spatially covariant lesion segmentation},
  author={Zhang, Hang and Wang, Rongguang and Zhang, Jinwei and Liu, Dongdong and Li, Chao and Li, Jiahao},
  booktitle={Proceedings of the Thirty-Second International Joint Conference on Artificial Intelligence},
  pages={1713--1721},
  year={2023}
}

@article{avants2008symmetric,
  title= {{Symmetric diffeomorphic image registration with cross-correlation: Evaluating automated labeling of elderly and neurodegenerative brain}},
  author={Avants, Brian B and Epstein, Charles L and Grossman, Murray and Gee, James C},
  journal={Medical Image Analysis},
  volume={12},
  number={1},
  pages={26--41},
  year={2008},
  publisher={Elsevier}
}

@article{su2009augmented,
  title={Augmented reality during robot-assisted laparoscopic partial nephrectomy: toward real-time 3D-CT to stereoscopic video registration},
  author={Su, Li-Ming and Vagvolgyi, Balazs P and Agarwal, Rahul and Reiley, Carol E and Taylor, Russell H and Hager, Gregory D},
  journal={Urology},
  volume={73},
  number={4},
  pages={896--900},
  year={2009},
  publisher={Elsevier}
}

@inproceedings{chen2013voxel,
  title={Voxel-wise displacement as independent features in classification of multiple sclerosis},
  author={Chen, Min and Carass, Aaron and Reich, Daniel S and Calabresi, Peter A and Pham, Dzung and Prince, Jerry L},
  booktitle={Proceedings of SPIE},
  volume={8669},
  pages={10--1117},
  year={2013}
}

@article{fechter2020one,
  title={One-shot learning for deformable medical image registration and periodic motion tracking},
  author={Fechter, Tobias and Baltas, Dimos},
  journal={IEEE Transactions on Medical Imaging},
  volume={39},
  number={7},
  pages={2506--2517},
  year={2020},
  publisher={IEEE}
}

@article{chen2025lumir,
  title={Beyond the LUMIR challenge: The pathway to foundational registration models},
  author={Chen, Junyu and Wei, Shuwen and Honkamaa, Joel and Marttinen, Pekka and Zhang, Hang and Liu, Min and Zhou, Yichao and Tan, Zuopeng and Wang, Zhuoyuan and Wang, Yi and others},
  journal={arXiv preprint arXiv:2505.24160},
  year={2025}
}

@inproceedings{yang2024shiftmorph,
  title={ShiftMorph: A Fast and Robust Convolutional Neural Network for 3D Deformable Medical Image Registration},
  author={Yang, Lijian and Li, Weisheng and Shu, Yucheng and Mi, Jianxun and Huang, Yuping and Xiao, Bin},
  booktitle={ACM Multimedia},
  year={2024}
}

@inproceedings{meng2024correlation,
  title={Correlation-aware Coarse-to-fine MLPs for Deformable Medical Image Registration},
  author={Meng, Mingyuan and Feng, Dagan and Bi, Lei and Kim, Jinman},
  booktitle={Proceedings of the IEEE/CVF Conference on Computer Vision and Pattern Recognition},
  pages={9645--9654},
  year={2024}
}

@inproceedings{ma2024iirp,
  title={IIRP-Net: Iterative Inference Residual Pyramid Network for Enhanced Image Registration},
  author={Ma, Tai and Zhang, Suwei and Li, Jiafeng and Wen, Ying},
  booktitle={Proceedings of the IEEE/CVF Conference on Computer Vision and Pattern Recognition},
  pages={11546--11555},
  year={2024}
}

@article{chen2022transmorph,
  title={Transmorph: Transformer for unsupervised medical image registration},
  author={Chen, Junyu and Frey, Eric C and He, Yufan and Segars, William P and Li, Ye and Du, Yong},
  journal={Medical Image Analysis},
  volume={82},
  pages={102615},
  year={2022},
  publisher={Elsevier}
}

@inproceedings{zhang2025voxelopt,
  title={VoxelOpt: Voxel-Adaptive Message Passing for Discrete Optimization in Deformable Abdominal CT Registration},
  author={Zhang, Hang and Zhang, Yuxi and Wang, Jiazheng and Chen, Xiang and Hu, Renjiu and Tian, Xin and Li, Gaolei and Liu, Min},
  booktitle={International Conference on Medical Image Computing and Computer-Assisted Intervention},
  pages={672--683},
  year={2025},
  organization={Springer}
}

@article{he2025samir,
  title={SAMIR, an efficient registration framework via robust feature learning from SAM},
  author={He, Yue and Liu, Min and Liu, Qinghao and Wang, Jiazheng and Wang, Yaonan and Zhang, Hang and Chen, Xiang},
  journal={arXiv preprint arXiv:2509.13629},
  year={2025}
}

@inproceedings{zhang2025unsupervised,
  title={Unsupervised deformable image registration with structural nonparametric smoothing},
  author={Zhang, Hang and Hu, Renjiu and Chen, Xiang and Liu, Min and Wang, Yaonan and Wang, Rongguang and Zhang, Jinwei and Li, Gaolei and Cheng, Xinxing and Duan, Jinming},
  booktitle={International Conference on Information Processing in Medical Imaging},
  pages={108--124},
  year={2025},
  organization={Springer}
}

@inproceedings{islam2019much,
  title={How much Position Information Do Convolutional Neural Networks Encode?},
  author={Islam, Md Amirul and Jia, Sen and Bruce, Neil DB},
  booktitle={International Conference on Learning Representations},
  year={2020}
}

@article{jia2024decoder,
  title={Decoder-only image registration},
  author={Jia, Xi and Lu, Wenqi and Cheng, Xinxing and Duan, Jinming},
  journal={IEEE Transactions on Medical Imaging},
  year={2025},
  publisher={IEEE}
}

@article{horn1981determining,
  title={Determining optical flow},
  author={Horn, Berthold KP and Schunck, Brian G},
  journal={Artificial Intelligence},
  volume={17},
  number={1-3},
  pages={185--203},
  year={1981},
  publisher={Elsevier}
}

@article{zhao2019unsupervised,
  title = {Unsupervised 3D End-to-End Medical Image Registration with Volume Tweening Network},
  author = {Zhao, Shengyu and Lau, Tingfung and Luo, Ji and Chang, Eric I and Xu, Yan},
  journal = {IEEE Journal of Biomedical and Health Informatics},
  year = {2019},
  doi = {10.1109/JBHI.2019.2951024}
}

@article{dufumier2022openbhb,
  title={Openbhb: a large-scale multi-site brain mri data-set for age prediction and debiasing},
  author={Dufumier, Benoit and Grigis, Antoine and Victor, Julie and Ambroise, Corentin and Frouin, Vincent and Duchesnay, Edouard},
  journal={NeuroImage},
  volume={263},
  pages={119637},
  year={2022},
  publisher={Elsevier}
}

@article{taha2023magnetic,
  title={Magnetic resonance imaging datasets with anatomical fiducials for quality control and registration},
  author={Taha, Alaa and Gilmore, Greydon and Abbass, Mohamad and Kai, Jason and Kuehn, Tristan and Demarco, John and Gupta, Geetika and Zajner, Chris and Cao, Daniel and Chevalier, Ryan and others},
  journal={Scientific Data},
  volume={10},
  number={1},
  pages={449},
  year={2023},
  publisher={Nature Publishing Group UK London}
}

@article{hoffmann2021synthmorph,
  title={SynthMorph: learning contrast-invariant registration without acquired images},
  author={Hoffmann, Malte and Billot, Benjamin and Greve, Douglas N and Iglesias, Juan Eugenio and Fischl, Bruce and Dalca, Adrian V},
  journal={IEEE Transactions on Medical Imaging},
  volume={41},
  number={3},
  pages={543--558},
  year={2021},
  publisher={IEEE}
}

@article{simpson2019large,
  title={A large annotated medical image dataset for the development and evaluation of segmentation algorithms},
  author={Simpson, Amber L and Antonelli, Michela and Bakas, Spyridon and Bilello, Michel and Farahani, Keyvan and Van Ginneken, Bram and Kopp-Schneider, Annette and Landman, Bennett A and Litjens, Geert and Menze, Bjoern and others},
  journal={arXiv preprint arXiv:1902.09063},
  year={2019}
}

@ARTICLE{wang_RDP,
  author={Wang, Haiqiao and Ni, Dong and Wang, Yi},
  journal={IEEE Transactions on Medical Imaging}, 
  title={Recursive Deformable Pyramid Network for Unsupervised Medical Image Registration}, 
  year={2024},
  volume={},
  number={},
  pages={1-1},
  doi={10.1109/TMI.2024.3362968}}

@article{luo2016understanding,
  title={Understanding the effective receptive field in deep convolutional neural networks},
  author={Luo, Wenjie and Li, Yujia and Urtasun, Raquel and Zemel, Richard},
  journal={Advances in Neural Information Processing Systems},
  volume={29},
  year={2016}
}

@article{bernard2018deep,
  title={{Deep Learning Techniques for Automatic MRI Cardiac Multi-structures Segmentation and Diagnosis: Is the Problem Solved?}},
  author={Bernard, Olivier and Lalande, Alain and Zotti, Clement and Cervenansky, Frederick and Yang, Xin and Heng, Pheng-Ann and Cetin, Irem and Lekadir, Karim and Camara, Oscar and Ballester, Miguel Angel Gonzalez and others},
  journal={IEEE Transactions on Medical Imaging},
  volume={37},
  number={11},
  pages={2514--2525},
  year={2018},
  publisher={ieee}
}

@article{wang2015quantitative,
  title={Quantitative susceptibility mapping (QSM): decoding MRI data for a tissue magnetic biomarker},
  author={Wang, Yi and Liu, Tian},
  journal={Magnetic Resonance in Medicine},
  volume={73},
  number={1},
  pages={82--101},
  year={2015},
  publisher={Wiley Online Library}
}

@article{balakrishnan2019voxelmorph,
  title={VoxelMorph: a learning framework for deformable medical image registration},
  author={Balakrishnan, Guha and Zhao, Amy and Sabuncu, Mert R and Guttag, John and Dalca, Adrian V},
  journal={IEEE Transactions on Medical Imaging},
  volume={38},
  number={8},
  pages={1788--1800},
  year={2019},
  publisher={IEEE}
}

@inproceedings{liu2021swin,
  title={Swin transformer: Hierarchical vision transformer using shifted windows},
  author={Liu, Ze and Lin, Yutong and Cao, Yue and Hu, Han and Wei, Yixuan and Zhang, Zheng and Lin, Stephen and Guo, Baining},
  booktitle={Proceedings of the IEEE/CVF International Conference on Computer Vision},
  pages={10012--10022},
  year={2021}
}

@inproceedings{dosovitskiyimage,
  title={An Image is Worth 16x16 Words: Transformers for Image Recognition at Scale},
  author={Dosovitskiy, Alexey and Beyer, Lucas and Kolesnikov, Alexander and Weissenborn, Dirk and Zhai, Xiaohua and Unterthiner, Thomas and Dehghani, Mostafa and Minderer, Matthias and Heigold, Georg and Gelly, Sylvain and others},
  booktitle={International Conference on Learning Representations},
  year={2020}
}

@inproceedings{jia2022u,
  title={U-net vs transformer: Is u-net outdated in medical image registration?},
  author={Jia, Xi and Bartlett, Joseph and Zhang, Tianyang and Lu, Wenqi and Qiu, Zhaowen and Duan, Jinming},
  booktitle={International Workshop on Machine Learning in Medical Imaging},
  pages={151--160},
  year={2022},
  organization={Springer}
}

@inproceedings{albu2016low,
  title={Low complexity image registration techniques based on integral projections},
  author={Albu, Felix},
  booktitle={2016 International Conference on Systems, Signals and Image Processing},
  pages={1--4},
  year={2016},
  organization={IEEE}
}

@inproceedings{uniGradICON2024MICCAI,
  title={unigradicon: A foundation model for medical image registration},
  author={Tian, Lin and Greer, Hastings and Kwitt, Roland and Vialard, Francois-Xavier and San Jos{\'e} Est{\'e}par, Ra{\'u}l and Bouix, Sylvain and Rushmore, Richard and Niethammer, Marc},
  booktitle={International Conference on Medical Image Computing and Computer-Assisted Intervention},
  pages={749--760},
  year={2024},
  organization={Springer}
}

@ARTICLE{2021sar_review,
  author={Feng, Ruitao and Shen, Huanfeng and Bai, Jianjun and Li, Xinghua},
  journal={IEEE Geoscience and Remote Sensing Magazine}, 
  title={Advances and Opportunities in Remote Sensing Image Geometric Registration: A systematic review of state-of-the-art approaches and future research directions}, 
  year={2021},
  volume={9},
  number={4},
  pages={120-142},
  doi={10.1109/MGRS.2021.3081763}}

@article{xiang2024two,
  title={Two-stage registration of SAR images with large distortion based on superpixel segmentation},
  author={Xiang, Deliang and Pan, Xiaoyu and Ding, Huaiyue and Cheng, Jianda and Sun, Xiaokun},
  journal={IEEE Transactions on Geoscience and Remote Sensing},
  volume={62},
  pages={1--15},
  year={2024},
  publisher={IEEE}
}

@inproceedings{han2023diffeomorphic,
  title={Diffeomorphic image registration with neural velocity field},
  author={Han, Kun and Sun, Shanlin and Yan, Xiangyi and You, Chenyu and Tang, Hao and Naushad, Junayed and Ma, Haoyu and Kong, Deying and Xie, Xiaohui},
  booktitle={Proceedings of the IEEE/CVF Winter Conference on Applications of Computer Vision},
  pages={1869--1879},
  year={2023}
}

@article{lyu2018super,
  title={Super-resolution MRI through deep learning},
  author={Lyu, Qing and You, Chenyu and Shan, Hongming and Wang, Ge},
  journal={arXiv preprint arXiv:1810.06776},
  year={2018}
}

@article{sun2024medical,
  title={Medical image registration via neural fields},
  author={Sun, Shanlin and Han, Kun and You, Chenyu and Tang, Hao and Kong, Deying and Naushad, Junayed and Yan, Xiangyi and Ma, Haoyu and Khosravi, Pooya and Duncan, James S and others},
  journal={Medical Image Analysis},
  volume={97},
  pages={103249},
  year={2024},
  publisher={Elsevier}
}

@inproceedings{zhang2022learning,
  title={Learning correspondences of cardiac motion from images using biomechanics-informed modeling},
  author={Zhang, Xiaoran and You, Chenyu and Ahn, Shawn and Zhuang, Juntang and Staib, Lawrence and Duncan, James},
  booktitle={International Workshop on Statistical Atlases and Computational Models of the Heart},
  pages={13--25},
  year={2022},
  organization={Springer}
}

@article{chen2025ideal,
  title={Ideal Registration? Segmentation is All You Need},
  author={Chen, Xiang and Zhang, Fengting and Liu, Qinghao and Liu, Min and Wu, Kun and Wang, Yaonan and Zhang, Hang},
  journal={arXiv preprint arXiv:2509.15784},
  year={2025}
}

@inproceedings{zhang2024heteroscedastic,
  title={Heteroscedastic uncertainty estimation framework for unsupervised registration},
  author={Zhang, Xiaoran and Pak, Daniel H and Ahn, Shawn S and Li, Xiaoxiao and You, Chenyu and Staib, Lawrence H and Sinusas, Albert J and Wong, Alex and Duncan, James S},
  booktitle={International Conference on Medical Image Computing and Computer-Assisted Intervention},
  pages={651--661},
  year={2024},
  organization={Springer}
}

@article{li2025edge,
  title={Edge-constrained temporal superpixel segmentation and graph-structured energy optimization for PolSAR change detection},
  author={Li, Nengcai and Xiang, Deliang and Ding, Huaiyue and Xie, Yuzhen and Su, Yi},
  journal={ISPRS Journal of Photogrammetry and Remote Sensing},
  volume={229},
  pages={49--64},
  year={2025},
  publisher={Elsevier}
}

@article{li2025multiscale,
  title={Multiscale adaptive PolSAR image superpixel generation based on local iterative clustering and polarimetric scattering features},
  author={Li, Nengcai and Xiang, Deliang and Sun, Xiaokun and Hu, Canbin and Su, Yi},
  journal={ISPRS Journal of Photogrammetry and Remote Sensing},
  volume={220},
  pages={307--322},
  year={2025},
  publisher={Elsevier}
}

@article{heinrich2013mrf,
  title={MRF-based deformable registration and ventilation estimation of lung CT},
  author={Heinrich, Mattias P and Jenkinson, Mark and Brady, Michael and Schnabel, Julia A},
  journal={IEEE Transactions on Medical Imaging},
  volume={32},
  number={7},
  pages={1239--1248},
  year={2013},
  publisher={IEEE}
}

@article{lytle2020neuroimaging,
  title={A neuroimaging dataset on working memory and reward processing in children with and without ADHD},
  author={Lytle, Marisa N and Hammer, Rubi and Booth, James R},
  journal={Data in Brief},
  volume={31},
  pages={105801},
  year={2020},
  publisher={Elsevier}
}

@article{lytle2021neuroimaging,
  title={A neuroimaging dataset on response inhibition and selective attention in adults and children with and without ADHD},
  author={Lytle, Marisa N and Burman, Douglas D and Booth, James R},
  journal={Data in Brief},
  volume={37},
  pages={107158},
  year={2021},
  publisher={Elsevier}
}

@article{nugent2022nimh,
  title={The NIMH intramural healthy volunteer dataset: A comprehensive MEG, MRI, and behavioral resource},
  author={Nugent, Allison C and Thomas, Adam G and Mahoney, Margaret and Gibbons, Alison and Smith, Jarrod T and Charles, Antoinette J and Shaw, Jacob S and Stout, Jeffrey D and Namyst, Anna M and Basavaraj, Arshitha and others},
  journal={Scientific Data},
  volume={9},
  number={1},
  pages={518},
  year={2022},
  publisher={Nature Publishing Group UK London}
}

@article{mahler2024ultracortex,
  title={UltraCortex: Submillimeter ultra-high field 9.4 t brain mr image collection and manual cortical segmentations},
  author={Mahler, Lucas and Steiglechner, Julius and Bender, Benjamin and Lindig, Tobias and Ramadan, Dana and Bause, Jonas and Birk, Florian and Heule, Rahel and Charyasz, Edyta and Erb, Michael and others},
  journal={arXiv preprint arXiv:2406.18571},
  year={2024}
}

@inproceedings{chen2021deepDiscontinuity,
  title={A deep discontinuity-preserving image registration network},
  author={Chen, Xiang and Xia, Yan and Ravikumar, Nishant and Frangi, Alejandro F},
  booktitle={International Conference on Medical Image Computing and Computer Assisted Intervention},
  pages={46--55},
  year={2021},
  organization={Springer}
}

@article{hu2022cross,
  title={Cross-resolution distillation for efficient 3D medical image registration},
  author={Hu, Bo and Zhou, Shenglong and Xiong, Zhiwei and Wu, Feng},
  journal={IEEE Transactions on Circuits and Systems for Video Technology},
  volume={32},
  number={10},
  pages={7269--7283},
  year={2022},
  publisher={IEEE}
}

@article{wang2024improving,
  title={Improving misaligned multi-modality image fusion with one-stage progressive dense registration},
  author={Wang, Di and Liu, Jinyuan and Ma, Long and Liu, Risheng and Fan, Xin},
  journal={IEEE Transactions on Circuits and Systems for Video Technology},
  volume={34},
  number={11},
  pages={10944--10958},
  year={2024},
  publisher={IEEE}
}

@article{tang2022piafusion,
  title={PIAFusion: A progressive infrared and visible image fusion network based on illumination aware},
  author={Tang, Linfeng and Yuan, Jiteng and Zhang, Hao and Jiang, Xingyu and Ma, Jiayi},
  journal={Information Fusion},
  volume={83},
  pages={79--92},
  year={2022},
  publisher={Elsevier}
}

@article{zheng2025plug,
  title={Plug-and-Play General Image Registration for Misaligned Multi-Modal Image Fusion},
  author={Zheng, Tianheng and Dong, Guanglu and Zhang, Pingping and He, Xiaohai and Ren, Chao},
  journal={IEEE Transactions on Circuits and Systems for Video Technology},
  year={2025},
  publisher={IEEE}
}

@inproceedings{huang2022reconet,
  title={Reconet: Recurrent correction network for fast and efficient multi-modality image fusion},
  author={Huang, Zhanbo and Liu, Jinyuan and Fan, Xin and Liu, Risheng and Zhong, Wei and Luo, Zhongxuan},
  booktitle={European Conference on Computer Vision},
  pages={539--555},
  year={2022},
  organization={Springer}
}

@article{lowe2004distinctive,
  title={Distinctive image features from scale-invariant keypoints},
  author={Lowe, David G},
  journal={International Journal of Computer Vision},
  volume={60},
  number={2},
  pages={91--110},
  year={2004},
  publisher={Springer}
}

@article{tang2022superfusion,
  title={SuperFusion: A versatile image registration and fusion network with semantic awareness},
  author={Tang, Linfeng and Deng, Yuxin and Ma, Yong and Huang, Jun and Ma, Jiayi},
  journal={IEEE/CAA Journal of Automatica Sinica},
  volume={9},
  number={12},
  pages={2121--2137},
  year={2022},
  publisher={IEEE}
}

@inproceedings{zhao2019recursive,
  title={Recursive cascaded networks for unsupervised medical image registration},
  author={Zhao, Shengyu and Dong, Yue and Chang, Eric I and Xu, Yan and others},
  booktitle={Proceedings of the IEEE/CVF International Conference on Computer Vision},
  pages={10600--10610},
  year={2019}
}

@inproceedings{jia2023fourier,
  title={Fourier-net: Fast image registration with band-limited deformation},
  author={Jia, Xi and Bartlett, Joseph and Chen, Wei and Song, Siyang and Zhang, Tianyang and Cheng, Xinxing and Lu, Wenqi and Qiu, Zhaowen and Duan, Jinming},
  booktitle={Proceedings of the AAAI Conference on Artificial Intelligence},
  volume={37},
  number={1},
  pages={1015--1023},
  year={2023}
}

@inproceedings{arsigny2006log,
  title={A log-euclidean framework for statistics on diffeomorphisms},
  author={Arsigny, Vincent and Commowick, Olivier and Pennec, Xavier and Ayache, Nicholas},
  booktitle={International Conference on Medical Image Computing and Computer-Assisted Intervention},
  pages={924--931},
  year={2006},
  organization={Springer}
}

@inproceedings{wang2020deepflash,
  title={Deepflash: An efficient network for learning-based medical image registration},
  author={Wang, Jian and Zhang, Miaomiao},
  booktitle={Proceedings of the IEEE/CVF Conference on Computer Vision and Pattern Recognition},
  pages={4444--4452},
  year={2020}
}

@article{liu2024finite,
  title={On finite difference jacobian computation in deformable image registration},
  author={Liu, Yihao and Chen, Junyu and Wei, Shuwen and Carass, Aaron and Prince, Jerry},
  journal={International Journal of Computer Vision},
  pages={1--11},
  year={2024},
  publisher={Springer}
}

@inproceedings{ghahremani2024h,
  title={H-ViT: A Hierarchical Vision Transformer for Deformable Image Registration},
  author={Ghahremani, Morteza and Khateri, Mohammad and Jian, Bailiang and Wiestler, Benedikt and Adeli, Ehsan and Wachinger, Christian},
  booktitle={Proceedings of the IEEE/CVF Conference on Computer Vision and Pattern Recognition},
  pages={11513--11523},
  year={2024}
}

@article{chen2024textscf,
  title={Spatially covariant image registration with text prompts},
  author={Chen, Xiang and Liu, Min and Wang, Rongguang and Hu, Renjiu and Liu, Dongdong and Li, Gaolei and Zhang, Hang },
  journal={IEEE Transactions on Neural Networks and Learning Systems},
  pages={1-11},
  year={2024}
}

@inproceedings{li2023samconvex,
  title={SAMConvex: Fast Discrete Optimization for CT Registration Using Self-supervised Anatomical Embedding and Correlation Pyramid},
  author={Li, Zi and Tian, Lin and Mok, Tony CW and Bai, Xiaoyu and Wang, Puyang and Ge, Jia and Zhou, Jingren and Lu, Le and Ye, Xianghua and Yan, Ke and others},
  booktitle={International Conference on Medical Image Computing and Computer-Assisted Intervention},
  pages={559--569},
  year={2023},
  organization={Springer}
}

@inproceedings{dalca2018unsupervised,
  title={Unsupervised learning for fast probabilistic diffeomorphic registration},
  author={Dalca, Adrian V and Balakrishnan, Guha and Guttag, John and Sabuncu, Mert R},
  booktitle={International Conference on Medical Image Computing and Computer Assisted Intervention},
  pages={729--738},
  year={2018},
  organization={Springer}
}

@inproceedings{mok2020large,
  title={Large deformation diffeomorphic image registration with laplacian pyramid networks},
  author={Mok, Tony CW and Chung, Albert CS},
  booktitle={International Conference on Medical Image Computing and Computer Assisted Intervention},
  pages={211--221},
  year={2020},
  organization={Springer}
}

@article{kang2022dual,
  title={Dual-stream pyramid registration network},
  author={Kang, Miao and Hu, Xiaojun and Huang, Weilin and Scott, Matthew R and Reyes, Mauricio},
  journal={Medical Image Analysis},
  volume={78},
  pages={102379},
  year={2022},
  publisher={Elsevier}
}

@inproceedings{zhang2024memwarp,
  title={MemWarp: Discontinuity-Preserving Cardiac Registration with Memorized Anatomical Filters},
  author={Zhang, Hang and Chen, Xiang and Hu, Renjiu and Liu, Dongdong and Li, Gaolei and Wang, Rongguang},
  booktitle={International Conference on Medical Image Computing and Computer-Assisted Intervention},
  pages={671--681},
  year={2024},
  organization={Springer}
}

@article{wang2023robust,
  title={A robust and interpretable deep learning framework for multi-modal registration via keypoints},
  author={Wang, Alan Q and Evan, M Yu and Dalca, Adrian V and Sabuncu, Mert R},
  journal={Medical Image Analysis},
  volume={90},
  pages={102962},
  year={2023},
  publisher={Elsevier}
}

@inproceedings{ronneberger2015u,
  title={U-net: Convolutional networks for biomedical image segmentation},
  author={Ronneberger, Olaf and Fischer, Philipp and Brox, Thomas},
  booktitle={International Conference on Medical Image Computing and Computer-Assisted Intervention},
  pages={234--241},
  year={2015},
  organization={Springer}
}

@article{dalca2019unsupervised,
  title={{Unsupervised Learning of Probabilistic Diffeomorphic Registration for Images and Surfaces}},
  author={Dalca, Adrian V and Balakrishnan, Guha and Guttag, John and Sabuncu, Mert R},
  journal={Medical Image Analysis},
  volume={57},
  pages={226--236},
  year={2019},
  publisher={Elsevier}
}

@article{vercauteren2009diffeomorphic,
  title={Diffeomorphic demons: Efficient non-parametric image registration},
  author={Vercauteren, Tom and Pennec, Xavier and Perchant, Aymeric and Ayache, Nicholas},
  journal={NeuroImage},
  volume={45},
  number={1},
  pages={S61--S72},
  year={2009},
  publisher={Elsevier}
}

@article{islam2020much,
  title={How much position information do convolutional neural networks encode?},
  author={Islam, Md Amirul and Jia, Sen and Bruce, Neil DB},
  journal={arXiv preprint arXiv:2001.08248},
  year={2020}
}

@article{ashburner2007fast,
  title={A fast diffeomorphic image registration algorithm},
  author={Ashburner, John},
  journal={Neuroimage},
  volume={38},
  number={1},
  pages={95--113},
  year={2007},
  publisher={Elsevier}
}

@article{pratt1969hadamard,
  title={Hadamard transform image coding},
  author={Pratt, William K and Kane, Julius and Andrews, Harry C},
  journal={Proceedings of the IEEE},
  volume={57},
  number={1},
  pages={58--68},
  year={1969},
  publisher={IEEE}
}

@article{hugo2017longitudinal,
  title={A longitudinal four-dimensional computed tomography and cone beam computed tomography dataset for image-guided radiation therapy research in lung cancer},
  author={Hugo, Geoffrey D and Weiss, Elisabeth and Sleeman, William C and Balik, Salim and Keall, Paul J and Lu, Jun and Williamson, Jeffrey F},
  journal={Medical Physics},
  volume={44},
  number={2},
  pages={762--771},
  year={2017},
  publisher={Wiley Online Library}
}

@article{ulyanov2016instance,
  title={Instance normalization: The missing ingredient for fast stylization},
  author={Ulyanov, D},
  journal={arXiv preprint arXiv:1607.08022},
  year={2016}
}

@article{zhang2024slicer,
  title={Slicer Networks},
  author={Zhang, Hang and Chen, Xiang and Wang, Rongguang and Hu, Renjiu and Liu, Dongdong and Li, Gaolei},
  journal={arXiv preprint arXiv:2401.09833},
  year={2024}
}

@inproceedings{sandler2018mobilenetv2,
  title={Mobilenetv2: Inverted residuals and linear bottlenecks},
  author={Sandler, Mark and Howard, Andrew and Zhu, Menglong and Zhmoginov, Andrey and Chen, Liang-Chieh},
  booktitle={Proceedings of the IEEE Conference on Computer Vision and Pattern Recognition},
  pages={4510--4520},
  year={2018}
}

@inproceedings{lee2015deeply,
  title={Deeply-supervised nets},
  author={Lee, Chen-Yu and Xie, Saining and Gallagher, Patrick and Zhang, Zhengyou and Tu, Zhuowen},
  booktitle={Artificial Intelligence and Statistics},
  pages={562--570},
  year={2015},
  organization={Pmlr}
}

@article{hering2022learn2reg,
  title={Learn2Reg: comprehensive multi-task medical image registration challenge, dataset and evaluation in the era of deep learning},
  author={Hering, Alessa and Hansen, Lasse and Mok, Tony CW and Chung, Albert CS and Siebert, Hanna and H{\"a}ger, Stephanie and Lange, Annkristin and Kuckertz, Sven and Heldmann, Stefan and Shao, Wei and others},
  journal={IEEE Transactions on Medical Imaging},
  volume={42},
  number={3},
  pages={697--712},
  year={2022},
  publisher={IEEE}
}

@inproceedings{mok2021large,
  title={Large deformation image registration with anatomy-aware Laplacian pyramid networks},
  author={Mok, Tony CW and Chung, Albert CS},
  booktitle={Segmentation, Classification, and Registration of Multi-modality Medical Imaging Data: MICCAI 2020 Challenges},
  pages={61--67},
  year={2021},
  organization={Springer}
}

@inproceedings{balakrishnan2018unsupervised,
  title={An unsupervised learning model for deformable medical image registration},
  author={Balakrishnan, Guha and Zhao, Amy and Sabuncu, Mert R and Guttag, John and Dalca, Adrian V},
  booktitle={Proceedings of the IEEE Conference on Computer Vision and Pattern Recognition},
  pages={9252--9260},
  year={2018}
}

@inproceedings{mok2020fast,
  title={Fast symmetric diffeomorphic image registration with convolutional neural networks},
  author={Mok, Tony CW and Chung, Albert},
  booktitle={Proceedings of the IEEE/CVF Conference on Computer Vision and Pattern Recognition},
  pages={4644--4653},
  year={2020}
}

@inproceedings{ding2022scaling,
  title={Scaling up your kernels to 31x31: Revisiting large kernel design in cnns},
  author={Ding, Xiaohan and Zhang, Xiangyu and Han, Jungong and Ding, Guiguang},
  booktitle={Proceedings of the IEEE/CVF Conference on Computer Vision and Pattern Recognition},
  pages={11963--11975},
  year={2022}
}

@inproceedings{hoopes2021hypermorph,
  title={Hypermorph: Amortized hyperparameter learning for image registration},
  author={Hoopes, Andrew and Hoffmann, Malte and Fischl, Bruce and Guttag, John and Dalca, Adrian V},
  booktitle={International Conference on Information Processing in Medical Imaging},
  pages={3--17},
  year={2021},
  organization={Springer}
}

@article{marcus2007open,
  title={Open Access Series of Imaging Studies (OASIS): cross-sectional MRI data in young, middle aged, nondemented, and demented older adults},
  author={Marcus, Daniel S and Wang, Tracy H and Parker, Jamie and Csernansky, John G and Morris, John C and Buckner, Randy L},
  journal={Journal of Cognitive Neuroscience},
  volume={19},
  number={9},
  pages={1498--1507},
  year={2007},
  publisher={MIT Press One Rogers Street, Cambridge, MA 02142-1209, USA journals-info~…}
}

@article{xu2016evaluation,
  title={Evaluation of six registration methods for the human abdomen on clinically acquired CT},
  author={Xu, Zhoubing and Lee, Christopher P and Heinrich, Mattias P and Modat, Marc and Rueckert, Daniel and Ourselin, Sebastien and Abramson, Richard G and Landman, Bennett A},
  journal={IEEE Transactions on Biomedical Engineering},
  volume={63},
  number={8},
  pages={1563--1572},
  year={2016},
  publisher={IEEE}
}

@article{jia2021learning,
  title={Learning a model-driven variational network for deformable image registration},
  author={Jia, Xi and Thorley, Alexander and Chen, Wei and Qiu, Huaqi and Shen, Linlin and Styles, Iain B and Chang, Hyung Jin and Leonardis, Ales and De Marvao, Antonio and O’Regan, Declan P and others},
  journal={IEEE Transactions on Medical Imaging},
  volume={41},
  number={1},
  pages={199--212},
  year={2021},
  publisher={IEEE}
}

@inproceedings{siebert2021fast,
  title={Fast 3D registration with accurate optimisation and little learning for Learn2Reg 2021},
  author={Siebert, Hanna and Hansen, Lasse and Heinrich, Mattias P},
  booktitle={International Conference on Medical Image Computing and Computer-Assisted Intervention},
  pages={174--179},
  year={2021},
  organization={Springer}
}

@article{chen2021deep,
  title={Deep learning in medical image registration},
  author={Chen, Xiang and Diaz-Pinto, Andres and Ravikumar, Nishant and Frangi, Alejandro F},
  journal={Progress in Biomedical Engineering},
  volume={3},
  number={1},
  pages={012003},
  year={2021},
  publisher={IOP Publishing}
}

@article{zhang2025uniself,
  title={UNISELF: A Unified Network with Instance Normalization and Self-Ensembled Lesion Fusion for Multiple Sclerosis Lesion Segmentation},
  author={Zhang, Jinwei and Zuo, Lianrui and Dewey, Blake E and Remedios, Samuel W and Liu, Yihao and Hays, Savannah P and Pham, Dzung L and Mowry, Ellen M and Newsome, Scott D and Calabresi, Peter A and others},
  journal={arXiv preprint arXiv:2508.03982},
  year={2025}
}

@inproceedings{zhang2025bi,
  title={Bi-directional MS lesion filling and synthesis using denoising diffusion implicit model-based lesion repainting},
  author={Zhang, Jinwei and Zuo, Lianrui and Liu, Yihao and Remedios, Samuel and Landman, Bennett A and Prince, Jerry L and Carass, Aaron},
  booktitle={Medical Imaging 2025: Image Processing},
  volume={13406},
  pages={217--223},
  year={2025},
  organization={SPIE}
}

@inproceedings{zhang2025aultra,
  title     = {AUTOMATED UNIQUE LESION TRACKING (AULTRA): A FRAMEWORK FOR LONGITUDINAL LESION-WISE MORPHOMETRY ANALYSIS IN MULTIPLE SCLEROSIS},
  author    = {Zhang, Jinwei and Rivas, Carlos and Dewey, Blake E. and Wei, Shuwen and Zhang, Hang and Remedios, Samuel W. and Hays, Savannah P. and Wang, Shimeng and Zuo, Lianrui and Mowry, Ellen M. and Newsome, Scott D. and Saidha, Shiv and Calabresi, Peter A. and Prince, Jerry L. and Carass, Aaron},
  booktitle = {Radiological Society of North America (RSNA) Annual Meeting},
  year      = {2025},
  
}

@inproceedings{rivas2025unique,
  title={Unique MS lesion identification from MRI},
  author={Rivas, Carlos A and Zhang, Jinwei and Wei, Shuwen and Remedios, Samuel W and Carass, Aaron and Prince, Jerry L},
  booktitle={Medical Imaging 2025: Image Processing},
  volume={13406},
  pages={592--599},
  year={2025},
  organization={SPIE}
}

\begin{IEEEbiography}[{\includegraphics[width=1in,height=1.25in,clip,keepaspectratio]{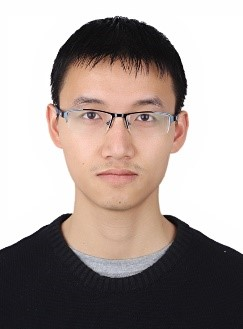}}]{Xiang Chen}
received his B.S. degree in Electronics and Information Engineering in 2016 and the M.S. degree in Communication and Information System in 2019, both from Sichuan University, Chengdu, China. He received his PhD degree from School of Computing, University of Leeds, Leeds, UK, in 2023. He is currently an assistant Professor with the College of Electrical and Information Engineering, Hunan University, Changsha, China. His research interests include computer vision and medical image analysis.
\end{IEEEbiography}

\begin{IEEEbiography}[{\includegraphics[width=1in,height=1.25in,clip,keepaspectratio]{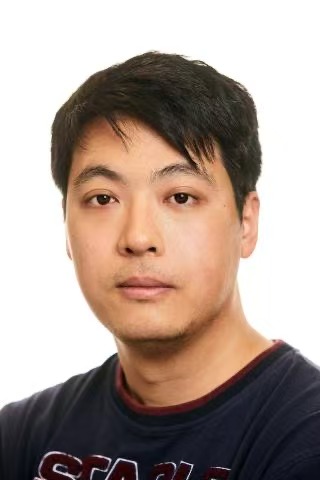}}]{Renjiu Hu}
received the B.S. degree in applied physics from University of Science and Technology of China, Hefei, Anhui, China in 2014 and the M.S. degree in Mechanical Engineering from Cornell University, Ithaca, NY, USA, in 2021. He is currently a PhD candidate in Mechanical Engineering from Cornell University, Ithaca, NY, USA. His current research interests include medical imaging analysis, machine learning, computer vision, and numerical simulation.
\end{IEEEbiography}

\begin{IEEEbiography}[{\includegraphics[width=1in,height=1.25in,clip,keepaspectratio]{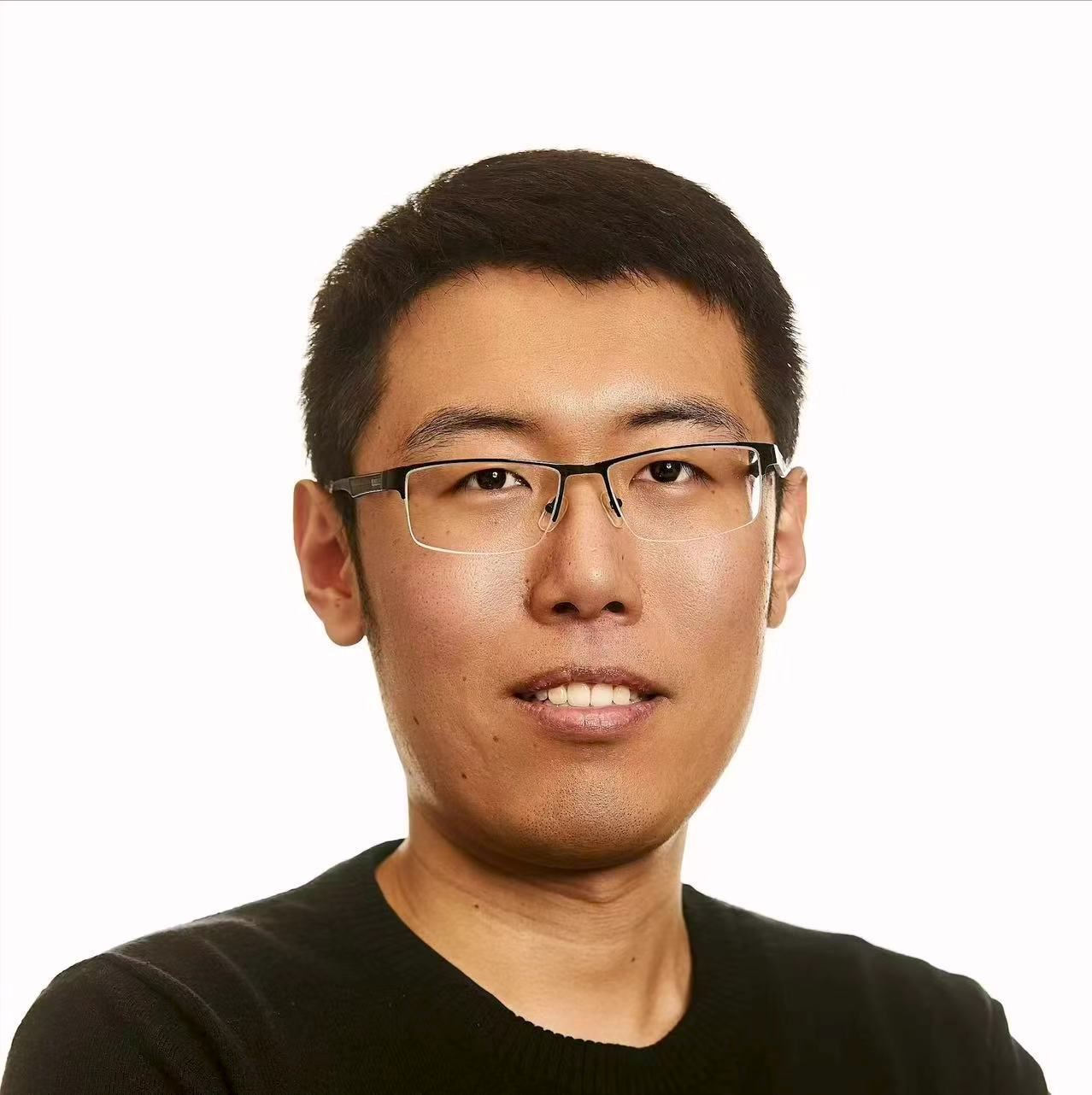}}]{Jinwei Zhang}
received the B.S. degree in Optical Information Science and Technology from Sun Yat-sen University, Guangzhou, China, in 2016, and a dual B.S. degree in Information and Computing Science from the same university in 2017. He received the Ph.D. degree in Biomedical Engineering from Cornell University, Ithaca, NY, USA, in 2023. He is currently a Postdoctoral Fellow in Electrical and Computer Engineering at Johns Hopkins University, USA. His research focuses on advanced magnetic resonance imaging (MRI) acquisition and reconstruction, quantitative susceptibility mapping, and lesion-wise analysis in neurological diseases including multiple sclerosis and Alzheimer’s disease. 
\end{IEEEbiography}

\begin{IEEEbiography}[{\includegraphics[width=1in,height=1.25in,clip,keepaspectratio]{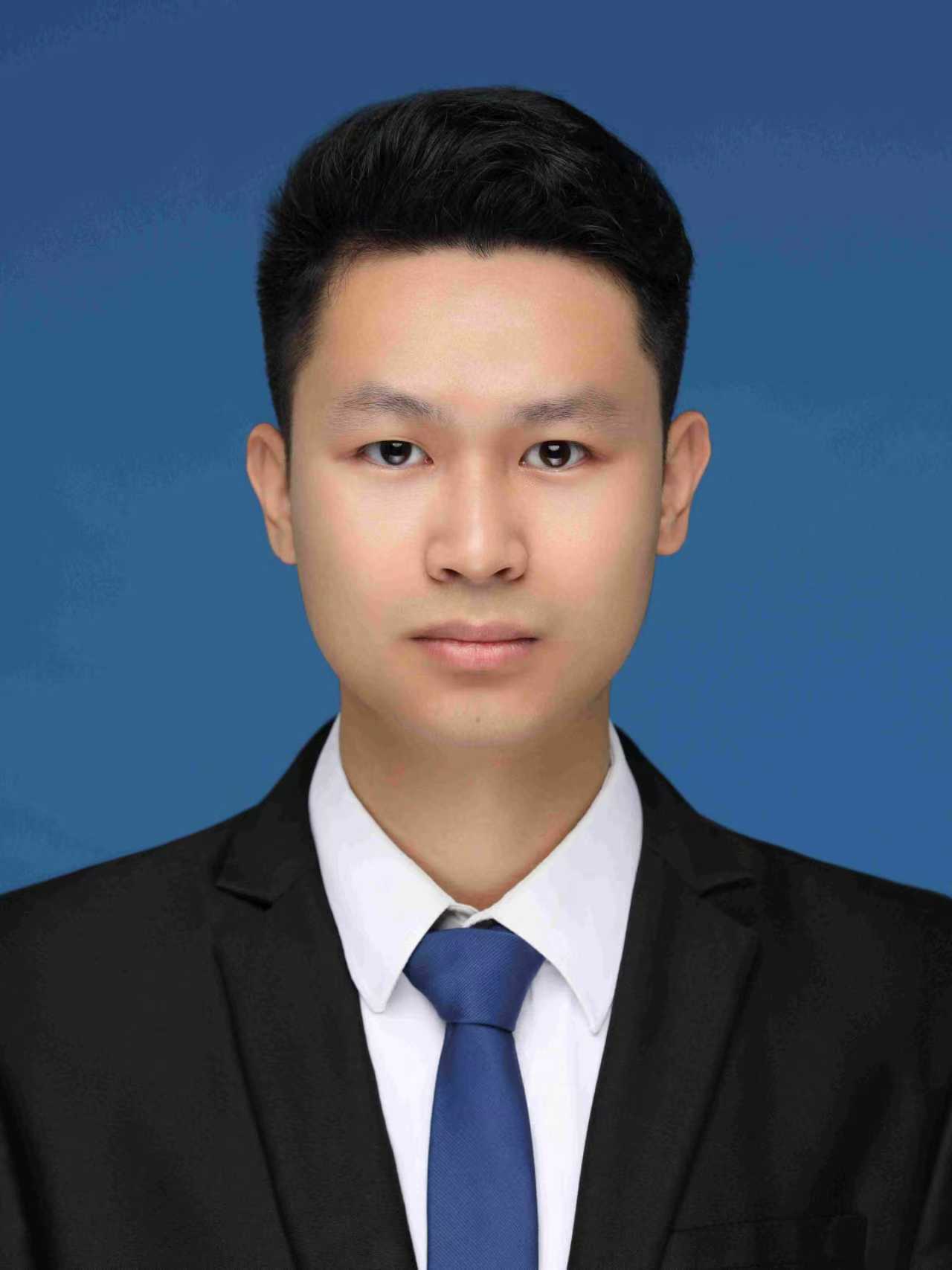}}]{Yuxi Zhang}
is a master student in the School of Artificial Intelligence and Robotics at Hunan University, China. His research interests include medical image registration and vision language model.
\end{IEEEbiography}

\begin{IEEEbiography}[{\includegraphics[width=1in,height=1.25in,clip,keepaspectratio]{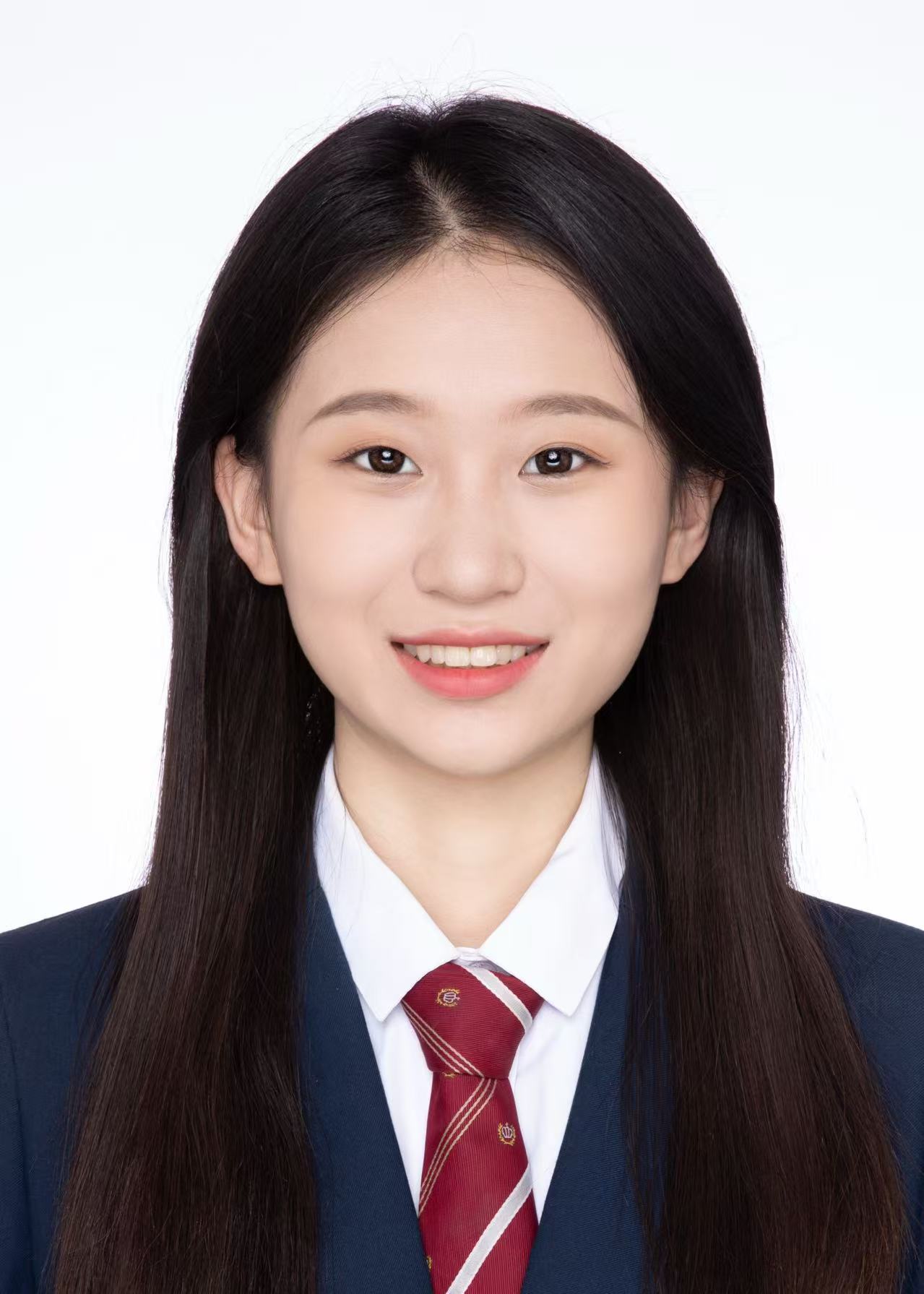}}]{Xinyao Yu}
received her B.S. degree in Internet of Things Engineering from Queen Mary University of London in 2023, and her M.S. degree in Computer Engineering from the National University of Singapore in 2025. Her current research interests include medical imaging analysis, computer vision, natural language processing, and intelligent sensing and communications.
\end{IEEEbiography}

\begin{IEEEbiography}[{\includegraphics[width=1in,height=1.25in,clip,keepaspectratio]{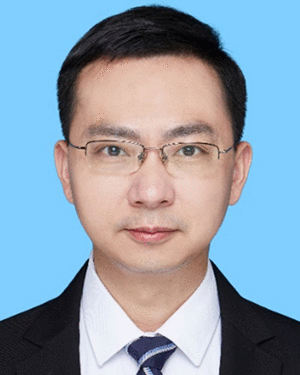}}]{Min Liu}
(Member, IEEE) received the bachelor’s degree from Beijing University, Beijing, China, in 2004, and the Ph.D. degree in electrical engineering from the University of California at Riverside, Riverside, CA, USA, in 2012. He is a Professor with the College of Electrical and Information Engineering, Hunan University, Changsha, China. He was a Research Scientist with the University of California at Santa Barbara, Santa Barbara, CA, USA. His research interests include computer vision and image processing. Dr. Liu is an Associate Editor of the IEEE TRANSACTIONS ON NEURAL NETWORKS AND LEARNING SYSTEMS.
\end{IEEEbiography}

\begin{IEEEbiography}[{\includegraphics[width=1in,height=1.25in,clip,keepaspectratio]{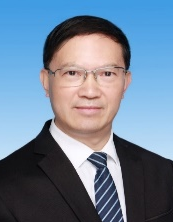}}]{Yaonan Wang}
received the Ph.D. degree in electrical engineering from Hunan University, Changsha, China, in 1994. Since 1995, he has been a Professor with the College of Electrical and Information Engineering, Hunan University. From 1994 to 1995, he was a Post-Doctoral Research Fellow with the Normal University of Defense Technology, Changsha. From 1998 to 2000, he was supported as a Senior Humboldt Fellow by the Federal Republic of Germany, University of Bremen, Bremen, Germany. From 2001 to 2004, he was a Visiting Professor with the University of Bremen. His research interests include robotics and image processing. Prof. Wang is a member of the Chinese Academy of Engineering.
\end{IEEEbiography}

\begin{IEEEbiography}[{\includegraphics[width=1in,height=1.25in,clip,keepaspectratio]{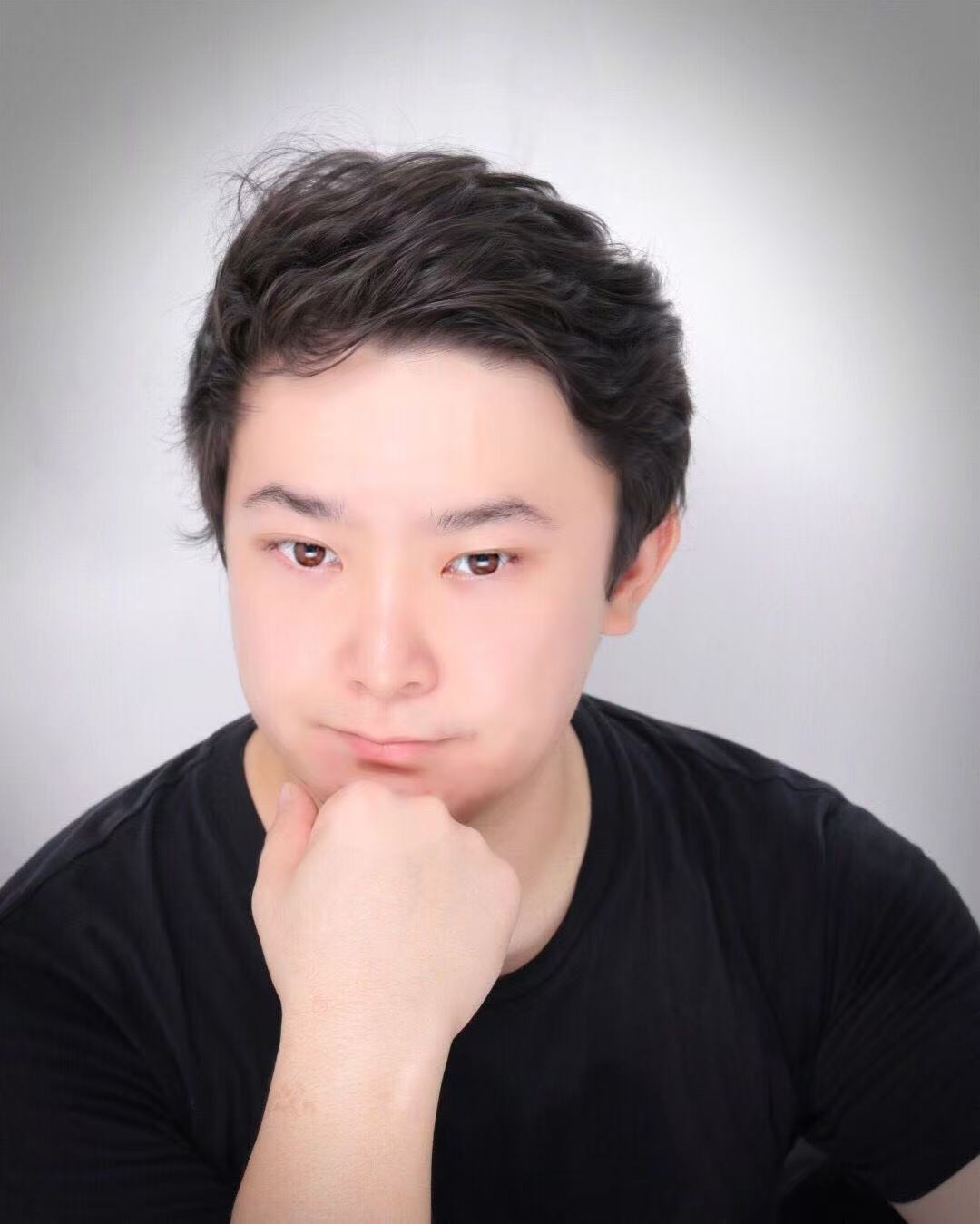}}]{Hang Zhang}
received his B.S. degree in Electronics and Information Engineering from Sichuan University in 2015, and his M.Phil. degree in Computer Science and Engineering from The Chinese University of Hong Kong in 2017. He earned his Ph.D. in Electrical and Computer Engineering from Cornell University. He has received two Best Paper Awards, at the International Symposium on Physical Design in 2017 and the Field-Programmable Custom Computing Machines conference in 2018. He was also nominated for a Best Paper Award at the International Symposium on Biomedical Imaging in 2021. His current research focuses on developing novel neural network models and theories for general-purpose foundation models, with applications in healthcare, computer vision and robotics.
\end{IEEEbiography}


%

\newpage
\section{Appendix}
\subsection{Heatmap Generation of Fig. \ref{fig:conv_vis}}
\label{sec:supp_heatmap} 
As part of our methodology for crafting the heatmap depicted in Fig. \ref{fig:conv_vis} of the main text, we began by training the EOIR network, with start channels $N_s=32$, on the abdomen CT dataset. 
Subsequently, we randomly selected an image from this dataset as the fixed image and derived the moving image by translating it one voxel anteriorly. Similarly, the three binary squares on the left are all translated with one voxel to the bottom to obtain the moving images, which is why their gradients respond to the vertical direction.

With the new moving and fixed images as inputs, we used the 3-layer encoder to extract a 32-channel feature map. 
To make the output visually interpretable, we applied Principal Component Analysis (PCA) to reduce the channels to one, retaining only the component with the largest variance. 
The heatmap was computed using the formula $(\mathbf{I}_{m}(p) - \mathbf{I}_{f}(p))/\frac{\partial \mathbf{I}_{f}}{\partial y}(p)$. 
Post-processing involved thresholding, where invalid voxels were set to 0 and valid voxels to 1. 
A 2D heatmap was then generated by averaging all slices along the axis direction and visualized using the `viridis' color map to enhance clarity and interpretation.
The untrained heatmap was generated following the same process as described above, but with the encoder initialized randomly.

\textcolor{resp}{\subsection{Training Details for Each Dataset}}
\textcolor{resp}{
During training, all networks were trained for 300 epochs on the Abdomen CT, LUMIR, ACDC, HippocampusMR, and RGB-IR datasets. An extended training period of 700 epochs was used for the OASIS dataset to achieve optimal performance. Unless otherwise specified, the number of pyramid layers $n$ and the scaling-and-squaring step $N_s$ were set to 5 and 32, respectively, and the integration step $m$ was fixed to 7 in all experiments.
The similarity loss and regularization weight were configured as follows for each dataset:
\begin{itemize}
    \item For the Abdomen CT, HippocampusMR, and OASIS datasets, we used a combination of NCC loss and Dice loss as the similarity measure, with a loss weighting ratio of $L_{\text{NCC}} : L_{\text{Dice}} : R = 1:1:1$, where $R$ denotes the smoothness regularization term.
    \item For the ACDC dataset, MSE loss was employed with a regularization weight $\lambda = 0.01$, corresponding to a ratio $L_{\text{MSE}} : R = 1:0.01$.
    \item On the LUMIR dataset, NCC loss was used with $\lambda = 5$ and a learning rate of $4 \times 10^{-4}$ ($L_{\text{NCC}} : R = 1:5$).
    \item For the RGB-IR dataset, we utilized a combination of L1 loss and perceptual loss (with features extracted using a VGG network) as the similarity objective, weighted as $L_1 : L_P : R = 1:1:1$. To handle the two modalities, separate encoders of identical architecture were used to extract features from the RGB and infrared images independently. During training, paired images were warped via random ``affine + deformable'' deformation fields, following the data augmentation strategy described in \cite{zheng2025plug,wang2024improving}.
\end{itemize}
 A summary of the experimental setups is provided in Table \ref{tab:exp_setup}, where NCC, Dice, L1, $L_P$, and $\mathcal{R}$ denote the normalized cross-correlation loss, Dice loss, L1 loss, perceptual loss, and smoothness regularization, respectively.}


\begin{table}[!ht]
\centering
\caption{Comprehensive Summary of Experimental Setup.}
\label{tab:exp_setup}
\begin{tabular}{p{0.25\linewidth} p{0.2\linewidth} p{0.4\linewidth}}
\toprule
\textbf{Training Paradigm} & \textbf{Dataset} & \textbf{Configuration \& Baselines} \\
\midrule
\multirow{4}{*}{\textbf{Unsupervised}}
& ACDC & \textbf{Loss:} NCC + $\mathcal{R}$ (1:1), \textbf{Baselines}: VoxelMorph, TransMorph, LKUNet, FourierNet, CorrMLP, RDP, MemWarp\\
& LUMIR & \textbf{Loss:} NCC + $\mathcal{R}$ (1:1), \textbf{Baselines}: VoxelMorph, TransMorph, SynthMorph, DeedsBCV \\
& RGB-IR & \textbf{Loss:} L1 + $L_p$ + $\mathcal{R}$ (1:1:0.01), \textbf{Baselines}: SIFT, ReCoNet, Superfusion, IMF, PGMR \\
& ThoraxCBCT & \textbf{Loss:} NCC + $\mathcal{R}$ (1:1), \textbf{Baselines}: VoxelMorph++, deeds \\
\cmidrule(lr){3-3}
\multirow{3}{*}{\textbf{Weakly-supervised}}
& Abdomen CT & \textbf{Loss:} NCC + Dice + $\mathcal{R}$ (1:1:1), \textbf{Baselines}: VoxelMorph, TransMorph, LKUNet, LapIRN, CorrMLP, RDP, MemWarp, LessNet, FourierNet, ConvexAdam, SAMConvex, VoxelOpt \\
& OASIS & \textbf{Loss:} NCC + Dice + $\mathcal{R}$ (1:1:1), \textbf{Baselines}: VoxelMorph, TransMorph, LKUNet, LapIRN, LessNet, FourierNet, ConvexAdam \\
& HippocampusMR & \textbf{Loss:} NCC + Dice + $\mathcal{R}$ (1:1:1), \textbf{Baselines}: VoxelMorph, TransMorph, LKUNet, CorrMLP, RDP, MemWarp, LessNet, FourierNet \\
\cmidrule(lr){3-3}
\multirow{3}{*}{\textbf{Zero-Shot}}
& ADNI & \textbf{Loss:} N/A, \textbf{Baselines:} All methods in the LUMIR challenge  \\
& NIMH & \textbf{Loss:} N/A, \textbf{Baselines:} All methods in the LUMIR challenge  \\
& UltraCortex & \textbf{Loss:} N/A, \textbf{Baselines:} All methods in the LUMIR challenge \\
\bottomrule
\end{tabular}
\end{table}

\subsection{Ablation Study on the Encoder Design}
\label{sec:supp_encoder}
A detailed analysis of how the number of convolution layers $n_c$ in encoder affects registration performance is presented in Table~\ref{tab:number_conv}.
`EOIR-1CONV' to `EOIR-6CONV' represent models with $1 \sim 6$ $3 \times 3 \times 3$ convolution layers in the encoder. 
`EOIR-0CONV' uses a untrainable $1 \times 1 \times 1$ convolution to expand the input channel dimension to match the start channel size, effectively functioning like no convolution, as it barely affect local details.
`EOIR-UNet' applies a traditional UNet in the encoder. 
The registration Dice improves with up to three convolution layers, but adding more than three yields diminishing returns. 
Using a UNet increases model complexity significantly, yet results in lower accuracy. 
Thus, three convolution layers strike an optimal balance between registration accuracy and model complexity.

\subsection{Boxplot Results in ACDC Dataset and Abdomen Image Dataset}
\begin{figure}[t]
    \centering
    \includegraphics[width=\linewidth]{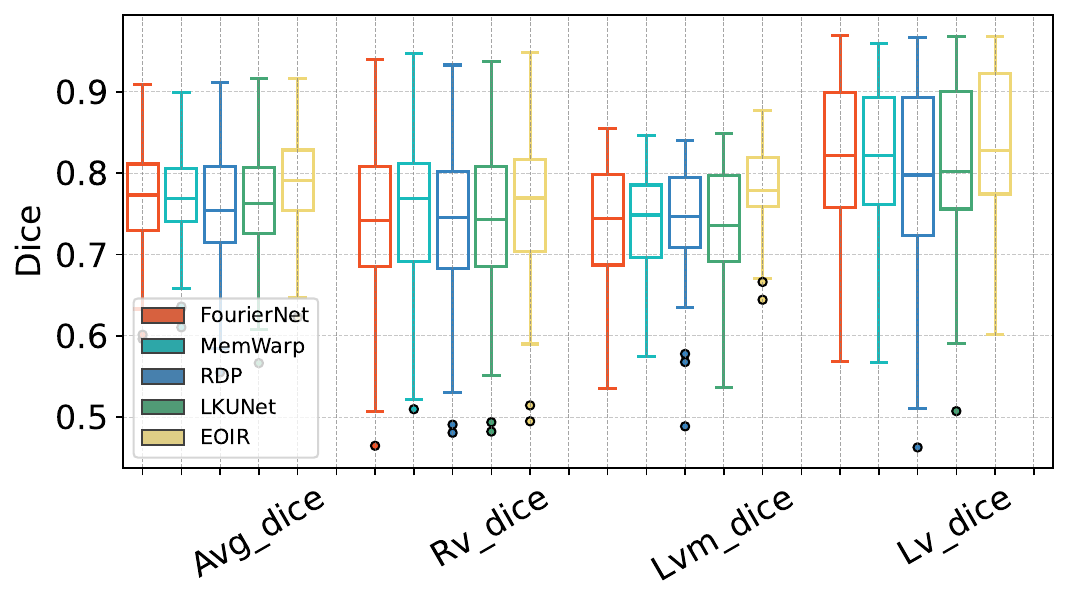}
    \caption{
        Boxplot results on cardiac MR image registration, where we compare our EOIR with MemWarp, FourierNet, LKUNet and RDP.
    }
    \label{fig:boxplot_acdc}
\end{figure}

\begin{figure*}[t]
    \centering
    \includegraphics[width=1.0\linewidth]{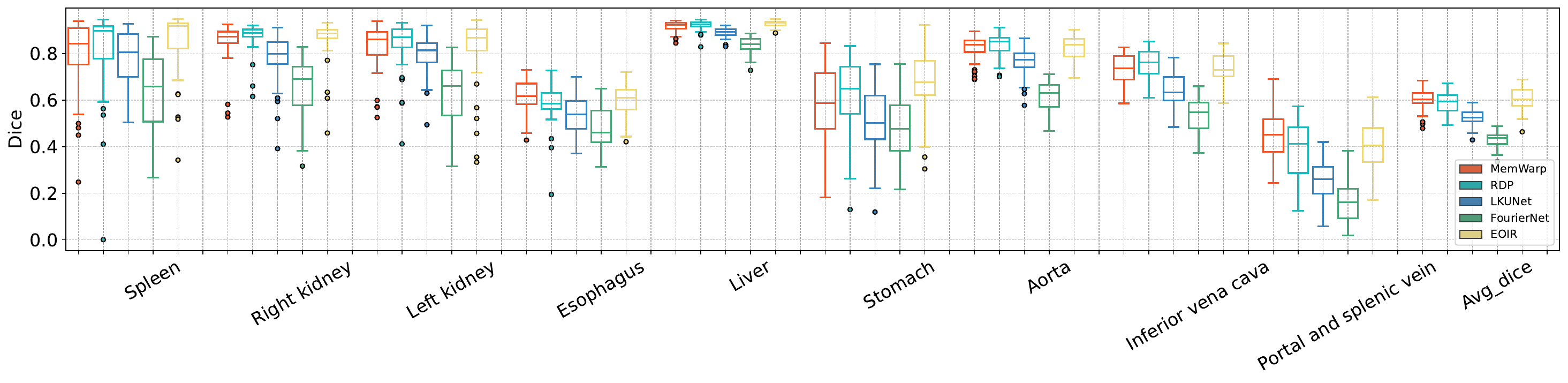}
    \caption{
        Boxplot results on abdominal image registration, where we compare our EOIR with MemWarp, FourierNet, LKUNet and RDP.
    }
    \label{fig:boxplot_abdomen}
\end{figure*}

The boxplot results in Figure~\ref{fig:boxplot_acdc} and Figure~\ref{fig:boxplot_abdomen} present the per-organ Dice and Avg. Dice scores of EOIR and the comparator methods. 
We compared EOIR against four top-performing methods: MemWarp, FourierNet, LKUNet, and RDP. 
In cardiac image registration, EOIR achieved a significantly higher Avg. Dice score than all other methods ($p<0.05$, t-test on Dice score). 
Similarly, in abdominal image registration, EOIR outperformed the other methods in Avg. Dice score ($p<0.05$), except for RDP ($p=0.099$).

\begin{figure}[t]
    \centering
    \includegraphics[width=1.0\linewidth]{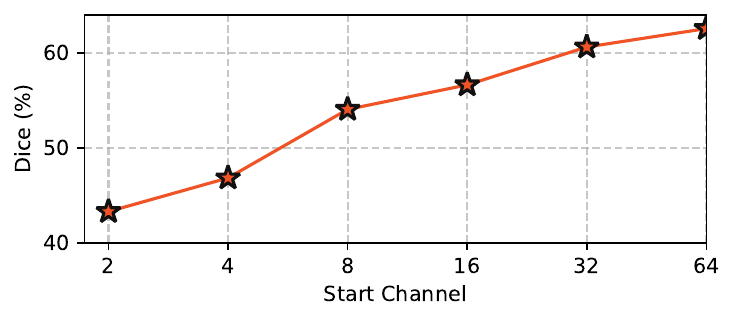}
    \caption{Dice scores of EOIR with the increasing start channels $Ns$ on abdomen image registration.
    }
    \label{fig:dice_vs_startchannel}
\end{figure}

\begin{table*}[!t]
\caption{
Quantitative comparison of registration results on the abdominal CT dataset, featuring different variations of EOIR. 
Metrics including Dice (\%), HD95 (mm), SDlogJ, multi-adds and Params are averaged across all image pairs for each method. 
Symbols indicate the desired direction of metric values: $\uparrow$ implies higher is better, while $\downarrow$ indicates lower is better. 
``Initial'' refers to the baseline results before registration.
}
\begin{center}
\resizebox{0.65\linewidth}{!}{
\begin{tabular}{ lccccc }
\hline
\hline
Model &  Dice (\%) $\uparrow$ & HD95 (mm) $\downarrow$ & SDlogJ $\downarrow$ & Multi-Adds (GB) $\downarrow$ & Params (MB) $\downarrow$\\ 
\hline
Initial & 30.86 & 29.77 & - & -&-\\
\hline
EOIR-0CONV  & 43.39 & 27.85 & 0.170& 179.35 &  0.80\\
EOIR-1CONV  & 54.40 & 21.26 & 0.172 & 180.98 & 0.80 \\
EOIR-2CONV \ & 55.92 & 20.92 & 0.177 & 235.40 & 0.83 \\
EOIR-3CONV  & 60.64 & 17.61 & 0.173 & 398.60 & 0.91 \\
EOIR-4CONV& 61.54 & 18.15 & 0.172 & 616.16& 1.02  \\
EOIR-5CONV & 62.23 & 17.17 & 0.170 & 833.72 & 1.13\\
EOIR-6CONV & 62.58 & 17.23 & 0.170 & 1050.00 & 1.24\\
EOIR-UNet & 58.27 & 17.22 & 0.168 & 1300.00 &  15.25\\
\hline
\hline
\end{tabular}
}
\end{center}
\label{tab:number_conv}
\end{table*}

\subsection{Effects of Varying Start Channels} 
We investigate the impact of varying the number of start channels \(N_s\) on registration performance.  
As shown in Figure~\ref{fig:dice_vs_startchannel}, Dice scores increase with \(N_s\) initially but plateau after \(N_s = 32\), accompanied by an exponential rise in model parameters.  
Thus, we set \(N_s = 32\) for most experiments in this work.

\subsection{LUMIR Ranking on Test Phase}
In the test phase of the LUMIR challenge, we still won the second ranking, with the detailed top three results in Table.~\ref{tab:lumir_ranking}. It can be observed that our EOIR achieves similar registration accuracy while keeping a significantly smoother deformation field, compared to the rest approaches.

\begin{table*}[!t]
\caption{
Quantitative results on the LUMIR dataset test phase, comparing EOIR with baseline methods and other participating teams.  
}
\begin{center}
\resizebox{0.65\linewidth}{!}{
\begin{tabular}{ lcccccc }
\hline
\hline
Team &  TRE\_LM $\downarrow$ & DSC (\%) $\uparrow$ & HD95 (mm) $\downarrow$ & NDV $\downarrow$ & Rank Score $\uparrow$ & Rank $\uparrow$\\ 
\hline
Initial & 4.38 & 0.55 & 4.91 & - & - & -\\
\hline
honkamj & 3.09 & 0.79 & 3.04 & 0.0025 & 0.814 & 1\\
\textbf{hnuzyx\_next-gen-nn (ours)} & 3.12 & 0.78 & 3.28 & 0.0001 & 0.781 & 2\\
lieweaver & 3.07 & 0.78 & 3.29 & 0.0121 & 0.737 & 3\\
uniGradICONwlO50 & 3.14 & 0.76 & 3.40 & 0.0002 & 0.668 & 7\\
VFA & 3.14 & 0.78 & 3.15 & 0.0704 & 0.667 & 8\\
TransMorph & 3.14 & 0.76 & 3.46 & 0.3621 & 0.518 & 11\\
deedsBCV & 3.10 & 0.70 & 3.94 & 0.0002 & 0.423 & 13\\
uniGradICON & 3.24 & 0.74 & 3.57 & 0.0001 & 0.402 & 14\\
SynthMorph & 3.23 & 0.72 & 3.61 & 0.0000 & 0.361 & 17\\
ANTsSyN & 3.48 & 0.70 & 3.69 & 0.0000 & 0.265 & 19\\
VoxelMorph & 3.53 & 0.71 & 4.07 & 1.2167 & 0.157 & 20\\
\hline
\hline
\end{tabular}
}
\end{center}
\label{tab:lumir_ranking}
\end{table*}

\subsection{Results of Zero-shot Inference}
The zero-shot inference capability of EOIR was validated on subject-to-atlas registrations using three distinct datasets (ADNI, NIMH, and UltraCortex), with the models trained exclusively on the LUMIR dataset and applied without any fine-tuning. As detailed in \cite{chen2025lumir} and Tables \ref{tab:adhd_atlas}, \ref{tab:nimh_t1w_atlas}, and \ref{tab:ultracortex_atlas} (where EOIR is listed as team `next-gen-nn'), our method consistently ranked among the top three in registration accuracy across all benchmarks while producing significantly smoother deformations than those approaches with comparable registration accuracy.

\begin{table*}[htbp]
\centering
\caption{Subject to atlas registration on ADHD dataset.}
\label{tab:adhd_atlas}
\resizebox{\textwidth}{!}{%
\begin{tabular}{rccccc}
\hline
 & \textbf{DSC} & \textbf{HD95} & \textbf{Ranking} & \textbf{NDV} & \textbf{DSC30} \\
\cmidrule(lr){2-2}\cmidrule(lr){3-3}\cmidrule(lr){4-4}\cmidrule(lr){5-5}\cmidrule(lr){6-6}
\textbf{Method} & \textbf{Mean($\pm$Std. Dev.)} & \textbf{Mean($\pm$Std. Dev.)} & \textbf{(ACC)} & \textbf{Mean($\pm$Std. Dev.)} & \textbf{Mean($\pm$Std. Dev.)} \\
\hline
\rowcolor{white}
Bailiang \sq{bailiang} & 0.774($\pm$0.011) & 3.127($\pm$0.385) & \rankcircle{1} & 7.61e-03($\pm$2.46e-03) & 0.762($\pm$0.007) \\
\rowcolor{tablerowgray}
next-gen-nn \sq{nextgennn} & 0.772($\pm$0.011) & 3.130($\pm$0.383) & \rankcircle{2} & 1.73e-04($\pm$1.64e-04) & 0.758($\pm$0.006) \\
\rowcolor{white}
honkamj \sq{honkamj} & 0.762($\pm$0.012) & 3.119($\pm$0.365) & \rankcircle{5} & 1.72e-03($\pm$2.73e-04) & 0.749($\pm$0.006) \\
\rowcolor{tablerowgray}
LoRA-FT \sq{loraft} & 0.741($\pm$0.016) & 3.441($\pm$0.423) & \rankcircle{15} & 5.95e-04($\pm$3.25e-04) & 0.724($\pm$0.009) \\
\rowcolor{white}
MadeForLife \sq{madeforlife} & 0.767($\pm$0.012) & 3.188($\pm$0.390) & \rankcircle{6} & 4.06e-03($\pm$2.09e-03) & 0.753($\pm$0.007) \\
\rowcolor{tablerowgray}
lukasf \sq{lukasf} & 0.763($\pm$0.012) & 3.246($\pm$0.396) & \rankcircle{8} & 5.43e-02($\pm$1.22e-02) & 0.750($\pm$0.007) \\
\rowcolor{white}
LYU1 \sq{lyu} & 0.763($\pm$0.013) & 3.197($\pm$0.382) & \rankcircle{7} & 5.03e-03($\pm$7.45e-04) & 0.749($\pm$0.008) \\
\rowcolor{tablerowgray}
TimH \sq{timh} & 0.719($\pm$0.014) & 3.530($\pm$0.393) & \rankcircle{18} & 0.00e+00($\pm$0.00e+00) & 0.704($\pm$0.008) \\
\rowcolor{white}
VROC \sq{vroc} & 0.706($\pm$0.010) & 3.781($\pm$0.369) & \rankcircle{20} & 9.87e-02($\pm$3.20e-02) & 0.695($\pm$0.006) \\
\rowcolor{tablerowgray}
DutchMasters \sq{dutchmasters} & 0.760($\pm$0.012) & 3.165($\pm$0.383) & \rankcircle{9} & 1.64e-03($\pm$1.02e-03) & 0.746($\pm$0.008) \\
\rowcolor{white}
zhuoyuanw210 \sq{zhuoyuan} & 0.766($\pm$0.012) & 3.125($\pm$0.371) & \rankcircle{3} & 1.30e-03($\pm$4.02e-04) & 0.752($\pm$0.007) \\
\hline
\rowcolor{tablerowgray}
ANTsSyN \sq{antssyn} & 0.745($\pm$0.013) & 3.282($\pm$0.393) & \rankcircle{12} & 0.00e+00($\pm$0.00e+00) & 0.730($\pm$0.007) \\
\rowcolor{white}
DeedsBCV \sq{deedsbcv} & 0.698($\pm$0.016) & 3.673($\pm$0.386) & \rankcircle{21} & 1.95e-04($\pm$4.43e-04) & 0.680($\pm$0.008) \\
\rowcolor{tablerowgray}
FireANTsGreedy \sq{fireantsgreedy} & 0.749($\pm$0.015) & 3.398($\pm$0.429) & \rankcircle{13} & 0.00e+00($\pm$0.00e+00) & 0.732($\pm$0.009) \\
\rowcolor{white}
FireANTsSyN \sq{fireantssyn} & 0.741($\pm$0.014) & 3.474($\pm$0.428) & \rankcircle{16} & 2.74e-05($\pm$2.05e-05) & 0.725($\pm$0.009) \\
\rowcolor{tablerowgray}
SynthMorph \sq{synthmorph} & 0.720($\pm$0.019) & 3.442($\pm$0.404) & \rankcircle{17} & 6.38e-06($\pm$6.73e-06) & 0.699($\pm$0.008) \\
\rowcolor{white}
TransMorph \sq{transmorph} & 0.762($\pm$0.012) & 3.244($\pm$0.390) & \rankcircle{10} & 1.08e-01($\pm$2.28e-02) & 0.748($\pm$0.007) \\
\rowcolor{tablerowgray}
uniGradICON \sq{unigradicon} & 0.740($\pm$0.013) & 3.379($\pm$0.392) & \rankcircle{14} & 1.51e-05($\pm$2.06e-05) & 0.726($\pm$0.008) \\
\rowcolor{white}
uniGradICONiso \sq{unigradiconiso} & 0.754($\pm$0.012) & 3.230($\pm$0.385) & \rankcircle{11} & 3.38e-05($\pm$7.31e-05) & 0.740($\pm$0.008) \\
\rowcolor{tablerowgray}
VFA \sq{vfa} & 0.763($\pm$0.014) & 3.075($\pm$0.375) & \rankcircle{4} & 8.12e-03($\pm$1.71e-03) & 0.748($\pm$0.006) \\
\rowcolor{white}
VoxelMorph \sq{voxelmorph} & 0.720($\pm$0.020) & 3.773($\pm$0.482) & \rankcircle{19} & 4.86e-01($\pm$7.56e-02) & 0.698($\pm$0.012) \\
\rowcolor{tablerowgray}
ZeroDisplacement \sq{zerodisplacement} & 0.569($\pm$0.031) & 4.590($\pm$0.518) & \rankcircle{22} & 0.00e+00($\pm$0.00e+00) & 0.534($\pm$0.015) \\
\hline
\end{tabular}
}
\end{table*}

\begin{table*}[htbp]
\centering
\caption{Subject to atlas registration on NIMH dataset.}
\label{tab:nimh_t1w_atlas}
\resizebox{\textwidth}{!}{%
\begin{tabular}{rccccc}
\hline
 & \textbf{DSC} & \textbf{HD95} & \textbf{Ranking} & \textbf{NDV} & \textbf{DSC30} \\
\cmidrule(lr){2-2}\cmidrule(lr){3-3}\cmidrule(lr){4-4}\cmidrule(lr){5-5}\cmidrule(lr){6-6}
\textbf{Method} & \textbf{Mean($\pm$Std. Dev.)} & \textbf{Mean($\pm$Std. Dev.)} & \textbf{(ACC)} & \textbf{Mean($\pm$Std. Dev.)} & \textbf{Mean($\pm$Std. Dev.)} \\
\hline
\rowcolor{white}
Bailiang \sq{bailiang} & 0.813($\pm$0.008) & 2.487($\pm$0.189) & \rankcircle{5} & 8.25e-03($\pm$2.47e-03) & 0.803($\pm$0.006) \\
\rowcolor{tablerowgray}
next-gen-nn \sq{nextgennn} & 0.811($\pm$0.009) & 2.448($\pm$0.189) & \rankcircle{1} & 1.80e-04($\pm$1.59e-04) & 0.800($\pm$0.006) \\
\rowcolor{white}
honkamj \sq{honkamj} & 0.806($\pm$0.012) & 2.424($\pm$0.226) & \rankcircle{2} & 2.05e-03($\pm$4.09e-04) & 0.791($\pm$0.008) \\
\rowcolor{tablerowgray}
LoRA-FT \sq{loraft} & 0.777($\pm$0.011) & 2.709($\pm$0.282) & \rankcircle{16} & 1.15e-03($\pm$1.08e-03) & 0.764($\pm$0.009) \\
\rowcolor{white}
MadeForLife \sq{madeforlife} & 0.809($\pm$0.011) & 2.482($\pm$0.219) & \rankcircle{6} & 5.96e-03($\pm$1.66e-03) & 0.795($\pm$0.008) \\
\rowcolor{tablerowgray}
lukasf \sq{lukasf} & 0.796($\pm$0.008) & 2.596($\pm$0.208) & \rankcircle{11} & 6.69e-02($\pm$1.50e-02) & 0.786($\pm$0.006) \\
\rowcolor{white}
LYU1 \sq{lyu} & 0.806($\pm$0.012) & 2.476($\pm$0.221) & \rankcircle{7} & 5.72e-03($\pm$8.02e-04) & 0.790($\pm$0.009) \\
\rowcolor{tablerowgray}
TimH \sq{timh} & 0.760($\pm$0.010) & 2.844($\pm$0.211) & \rankcircle{17} & 0.00e+00($\pm$0.00e+00) & 0.747($\pm$0.007) \\
\rowcolor{white}
VROC \sq{vroc} & 0.714($\pm$0.020) & 3.370($\pm$0.256) & \rankcircle{21} & 6.39e-02($\pm$2.44e-02) & 0.691($\pm$0.015) \\
\rowcolor{tablerowgray}
DutchMasters \sq{dutchmasters} & 0.801($\pm$0.008) & 2.444($\pm$0.204) & \rankcircle{8} & 3.47e-03($\pm$1.31e-03) & 0.791($\pm$0.006) \\
\rowcolor{white}
zhuoyuanw210 \sq{zhuoyuan} & 0.809($\pm$0.011) & 2.440($\pm$0.208) & \rankcircle{3} & 2.06e-03($\pm$5.49e-04) & 0.795($\pm$0.008) \\
\hline
\rowcolor{tablerowgray}
ANTsSyN \sq{antssyn} & 0.784($\pm$0.015) & 2.598($\pm$0.224) & \rankcircle{12} & 0.00e+00($\pm$0.00e+00) & 0.770($\pm$0.019) \\
\rowcolor{white}
DeedsBCV \sq{deedsbcv} & 0.729($\pm$0.012) & 3.059($\pm$0.230) & \rankcircle{20} & 2.18e-04($\pm$6.37e-04) & 0.715($\pm$0.007) \\
\rowcolor{tablerowgray}
FireANTsGreedy \sq{fireantsgreedy} & 0.792($\pm$0.013) & 2.699($\pm$0.271) & \rankcircle{13} & 0.00e+00($\pm$0.00e+00) & 0.776($\pm$0.009) \\
\rowcolor{white}
FireANTsSyN \sq{fireantssyn} & 0.785($\pm$0.015) & 2.749($\pm$0.305) & \rankcircle{15} & 3.69e-05($\pm$2.26e-05) & 0.767($\pm$0.011) \\
\rowcolor{tablerowgray}
SynthMorph \sq{synthmorph} & 0.751($\pm$0.013) & 2.773($\pm$0.255) & \rankcircle{18} & 7.71e-06($\pm$1.03e-05) & 0.735($\pm$0.009) \\
\rowcolor{white}
TransMorph \sq{transmorph} & 0.803($\pm$0.010) & 2.550($\pm$0.212) & \rankcircle{9} & 1.19e-01($\pm$2.36e-02) & 0.791($\pm$0.007) \\
\rowcolor{tablerowgray}
uniGradICON \sq{unigradicon} & 0.780($\pm$0.009) & 2.628($\pm$0.231) & \rankcircle{14} & 3.14e-05($\pm$3.62e-05) & 0.770($\pm$0.006) \\
\rowcolor{white}
uniGradICONiso \sq{unigradiconiso} & 0.794($\pm$0.008) & 2.496($\pm$0.217) & \rankcircle{10} & 1.20e-04($\pm$1.27e-04) & 0.785($\pm$0.006) \\
\rowcolor{tablerowgray}
VFA \sq{vfa} & 0.805($\pm$0.013) & 2.425($\pm$0.209) & \rankcircle{4} & 9.40e-03($\pm$1.79e-03) & 0.788($\pm$0.008) \\
\rowcolor{white}
VoxelMorph \sq{voxelmorph} & 0.768($\pm$0.017) & 3.009($\pm$0.329) & \rankcircle{19} & 5.10e-01($\pm$8.16e-02) & 0.748($\pm$0.012) \\
\rowcolor{tablerowgray}
ZeroDisplacement \sq{zerodisplacement} & 0.596($\pm$0.022) & 3.835($\pm$0.307) & \rankcircle{22} & 0.00e+00($\pm$0.00e+00) & 0.570($\pm$0.015) \\
\hline
\end{tabular}
}
\end{table*}

\begin{table*}[htbp]
\centering
\caption{Subject to atlas registration on UltraCortex-9.4T dataset.}
\label{tab:ultracortex_atlas}
\resizebox{\textwidth}{!}{%
\begin{tabular}{rccccc}
\hline
 & \textbf{DSC} & \textbf{HD95} & \textbf{Ranking} & \textbf{NDV} & \textbf{DSC30} \\
\cmidrule(lr){2-2}\cmidrule(lr){3-3}\cmidrule(lr){4-4}\cmidrule(lr){5-5}\cmidrule(lr){6-6}
\textbf{Method} & \textbf{Mean($\pm$Std. Dev.)} & \textbf{Mean($\pm$Std. Dev.)} & \textbf{(ACC)} & \textbf{Mean($\pm$Std. Dev.)} & \textbf{Mean($\pm$Std. Dev.)} \\
\hline
\rowcolor{white}
Bailiang \sq{bailiang} & 0.783($\pm$0.032) & 2.911($\pm$0.626) & \rankcircle{7} & 1.23e-02($\pm$4.29e-03) & 0.747($\pm$0.038) \\
\rowcolor{tablerowgray}
next-gen-nn \sq{nextgennn} & 0.784($\pm$0.035) & 2.844($\pm$0.691) & \rankcircle{2} & 2.65e-04($\pm$1.91e-03) & 0.745($\pm$0.043) \\
\rowcolor{white}
honkamj \sq{honkamj} & 0.778($\pm$0.031) & 2.794($\pm$0.616) & \rankcircle{6} & 2.35e-03($\pm$4.15e-04) & 0.743($\pm$0.038) \\
\rowcolor{tablerowgray}
LoRA-FT \sq{loraft} & 0.736($\pm$0.031) & 3.168($\pm$0.697) & \rankcircle{16} & 4.31e-03($\pm$2.48e-03) & 0.699($\pm$0.033) \\
\rowcolor{white}
MadeForLife \sq{madeforlife} & 0.787($\pm$0.030) & 2.832($\pm$0.627) & \rankcircle{1} & 7.03e-03($\pm$1.82e-03) & 0.753($\pm$0.036) \\
\rowcolor{tablerowgray}
lukasf \sq{lukasf} & 0.764($\pm$0.034) & 2.963($\pm$0.624) & \rankcircle{10} & 1.01e-01($\pm$2.11e-02) & 0.725($\pm$0.039) \\
\rowcolor{white}
LYU1 \sq{lyu} & 0.781($\pm$0.031) & 2.846($\pm$0.653) & \rankcircle{5} & 8.19e-03($\pm$1.08e-03) & 0.746($\pm$0.037) \\
\rowcolor{tablerowgray}
TimH \sq{timh} & 0.735($\pm$0.033) & 3.182($\pm$0.642) & \rankcircle{17} & 0.00e+00($\pm$0.00e+00) & 0.696($\pm$0.038) \\
\rowcolor{white}
VROC \sq{vroc} & 0.694($\pm$0.026) & 3.635($\pm$0.650) & \rankcircle{21} & 9.02e-02($\pm$7.41e-02) & 0.663($\pm$0.022) \\
\rowcolor{tablerowgray}
DutchMasters \sq{dutchmasters} & 0.760($\pm$0.028) & 2.919($\pm$0.688) & \rankcircle{9} & 1.14e-02($\pm$6.84e-03) & 0.728($\pm$0.031) \\
\rowcolor{white}
zhuoyuanw210 \sq{zhuoyuan} & 0.781($\pm$0.031) & 2.855($\pm$0.668) & \rankcircle{4} & 2.35e-03($\pm$7.61e-04) & 0.745($\pm$0.037) \\
\hline
\rowcolor{tablerowgray}
ANTsSyN \sq{antssyn} & 0.756($\pm$0.034) & 2.949($\pm$0.627) & \rankcircle{11} & 0.00e+00($\pm$0.00e+00) & 0.716($\pm$0.041) \\
\rowcolor{white}
DeedsBCV \sq{deedsbcv} & 0.696($\pm$0.027) & 3.415($\pm$0.596) & \rankcircle{20} & 9.95e-05($\pm$1.84e-04)& 0.663($\pm$0.027) \\
\rowcolor{tablerowgray}
FireANTsGreedy \sq{fireantsgreedy} & 0.762($\pm$0.036) & 3.092($\pm$0.677) & \rankcircle{13} & 0.00e+00($\pm$0.00e+00) & 0.720($\pm$0.041) \\
\rowcolor{white}
FireANTsSyN \sq{fireantssyn} & 0.756($\pm$0.034) & 3.144($\pm$0.703) & \rankcircle{14} & 3.14e-05($\pm$1.85e-05) & 0.717($\pm$0.039) \\
\rowcolor{tablerowgray}
SynthMorph \sq{synthmorph} & 0.711($\pm$0.034) & 3.213($\pm$0.706) & \rankcircle{19} & 7.31e-06($\pm$1.03e-05)& 0.670($\pm$0.036) \\
\rowcolor{white}
TransMorph \sq{transmorph} & 0.776($\pm$0.035) & 2.912($\pm$0.649) & \rankcircle{8} & 1.61e-01($\pm$2.71e-02) & 0.735($\pm$0.041) \\
\rowcolor{tablerowgray}
uniGradICON \sq{unigradicon} & 0.738($\pm$0.031) & 3.149($\pm$0.743) & \rankcircle{15} & 8.53e-05($\pm$7.06e-05) & 0.702($\pm$0.034) \\
\rowcolor{white}
uniGradICONiso \sq{unigradiconiso} & 0.756($\pm$0.030) & 2.977($\pm$0.690) & \rankcircle{12} & 4.57e-04($\pm$7.50e-04) & 0.721($\pm$0.035) \\
\rowcolor{tablerowgray}
VFA \sq{vfa} & 0.782($\pm$0.030) & 2.763($\pm$0.596) & \rankcircle{3} & 1.22e-02($\pm$2.15e-03) & 0.750($\pm$0.037) \\
\rowcolor{white}
VoxelMorph \sq{voxelmorph} & 0.733($\pm$0.043) & 3.496($\pm$0.785) & \rankcircle{18} & 6.33e-01($\pm$1.43e-01) & 0.681($\pm$0.049) \\
\rowcolor{tablerowgray}
ZeroDisplacement \sq{zerodisplacement} & 0.568($\pm$0.045) & 4.272($\pm$0.772) & \rankcircle{22} & 0.00e+00($\pm$0.00e+00) & 0.513($\pm$0.043) \\
\hline
\end{tabular}
}
\end{table*}

\subsection{Results on ThoraxCBCT}
The performance of EOIR on the ThoraxCBCT dataset (Table \ref{tab:thoraxcbct_ranking}) highlights a key architectural consideration. For this challenging task—which involves aligning pre-therapeutic FBCT with interventional low-dose CBCT—the significant domain gap between modalities necessitates a more powerful feature extractor. We observed that the lightweight EOIR (3 CONV) variant was insufficient for mapping these disparate images into a common feature space. In contrast, an EOIR variant employing a U-Net encoder achieved substantially higher Dice scores, underscoring that encoder capacity is critical for robust performance in complex, multi-modal registration scenarios, even within our streamlined framework.

\begin{table*}[!t]
\caption{
 Quantitative results on the ThoraxCBCT dataset, obtained from the online leaderboard.
}
\begin{center}
\begin{tabular}{ lcccc }
\hline
\hline
Team &  Dice (\%)  $\uparrow$ & TRE(KP) (mm) $\downarrow$ & HD95 (mm) $\downarrow$ & SDlogJ $\downarrow$\\ 
\hline
Initial & 31.3 & 9.91 & 55.36 & - \\
\hline
VoxelMorph++ & 50.3 & 13.68 & 28.56 & 0.129 \\
deeds & 64.8 & 11.32 & 29.03 & 0.152\\
EOIR(3 CONV) & 45.4 & 13.12 & 53.49 & 0.115\\
EOIR(U-Net) & 56.2 & 14.23 & 41.49 & 0.228\\
\hline
\hline
\end{tabular}
\end{center}
\label{tab:thoraxcbct_ranking}
\end{table*}

\begin{figure*}[t]
    \centering
    \includegraphics[width=1.0\linewidth]{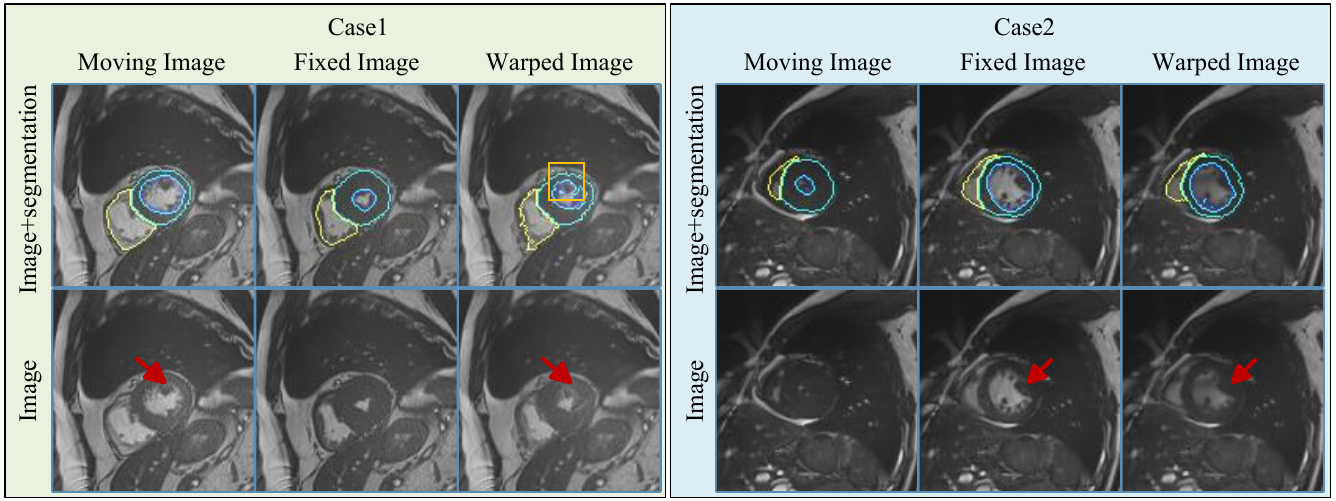}
    \caption{
        \textcolor{resp}{Two failure cases of EOIR on ACDC dataset (the worst results on ACDC). When there are inconsistent voxels in the moving and fixed images, our EOIR tend to produce unrealistic deformation due to align those local details.}
    }
    \label{fig:failure_case}
\end{figure*}

\textcolor{resp}{\subsection{Failure Case Analysis}}
\textcolor{resp}{To better understand the scenarios where EOIR may underperform, we visualize two of the worst registration results on the ACDC dataset. These failure cases are shown in Fig. \ref{fig:failure_case}. In the ACDC dataset, intra-subject registration is performed from end-diastole (ED) to end-systole (ES) and vice versa. Due to physiological changes between cardiac phases, some voxels in the moving image may lack direct correspondence in the fixed image. For instance, in Case 1, the small dark region in the moving image (highlighted with a red arrow) has no matching voxel in the fixed image. Similarly, in Case 2, certain areas in the fixed image are absent in the moving image (highlighted with a red arrow). In such cases, EOIR still attempts to establish correspondences, which can lead to locally unrealistic deformations (see the yellow box in Case 1). By contrast, in the ES-to-ED registration example, although structural inconsistencies remain, the resulting deformation appears more anatomically plausible. These results suggest that relying solely on voxel-level guidance may introduce locally implausible distortions. Therefore, incorporating anatomical shape priors could be essential for generating physically realistic deformations.
}

\vfill

\end{document}